\documentclass{article}

\usepackage[most]{tcolorbox}
\makeatletter
\def\blfootnote{\gdef\@thefnmark{}\@footnotetext}
\makeatletter

\usepackage[preprint]{corl_2026} 

\usepackage{graphicx} 
\usepackage{amsmath} 
\usepackage{booktabs}
\usepackage{multirow}
\usepackage{url}
\usepackage{makecell}
\usepackage{wrapfig}
\usepackage{float}
\usepackage{microtype}
\usepackage{enumitem}

\title{PIGEON: VLM-Driven Object Navigation via Points of Interest Selection}

\author{
  \textbf{Cheng Peng\textsuperscript{1,2,4,$*$}, Zhenzhe Zhang\textsuperscript{2,3,$*$}, Xiaobao Wei\textsuperscript{2,3}, Yanhao Zhang\textsuperscript{2}, Heng Wang\textsuperscript{2},} \\
  \textbf{Pengwei Wang\textsuperscript{2}, Zhongyuan Wang\textsuperscript{2}, Cheng Chi\textsuperscript{2,$\dagger$}, Shanghang Zhang\textsuperscript{2,3,$\dagger$}, Jing Liu\textsuperscript{1,4}} \\
  \textsuperscript{1}Institute of Automation, Chinese Academy of Sciences, \\
  \textsuperscript{2}Beijing Academy of Artificial Intelligence (BAAI), 
  \textsuperscript{3}Peking University \\
  \textsuperscript{4}School of Artificial Intelligence, University of Chinese Academy of Sciences
}
\begin{document}
\maketitle

\blfootnote{$^*$ Equal contribution. $^\dagger$ Corresponding author.}

\begin{abstract}
Object navigation in unseen indoor environments requires agents to perform semantic search under partial observability. Vision-language models (VLMs) provide strong semantic-spatial priors for this task, but how to interface them with robot navigation remains challenging: dense VLM inference is expensive, while abstracting environments into symbolic memories often separates high-level reasoning from the raw visual evidence that supports it. We propose \textit{\textbf{PIGEON}} (\textbf{P}oint of \textbf{I}nterest \textbf{G}uided \textbf{E}xploration for \textbf{O}bject \textbf{N}avigation), a VLM-driven framework that formulates object navigation as raw-observation-grounded sparse decision problem. PIGEON introduces Points of Interest (PoIs) as sparse visual decision units that couple geometrically executable waypoints with raw egocentric observations. Rather than using VLMs as dense controllers or restricting them to frontier ranking, PIGEON enables VLMs to select among task-critical PoIs, including exploration frontiers, suspected target objects, traversable stairs, and floor-level summaries, while low-level planners execute continuous motion between them. This PoI interface further makes high-level navigation decisions verifiable, allowing us to develop an RLVR pipeline that improves local VLMs without manual Chain-of-Thought annotations. Extensive experiments on Habitat ObjectNav benchmarks show that PIGEON achieves state-of-the-art zero-shot performance, scales consistently with foundation model capacity, and transfers to Active Embodied Question Answering with only prompt modifications. Real-world deployments on physical robots further demonstrate its robustness and efficiency.
\end{abstract}



	
    
\section{Introduction}

Object navigation in unseen indoor environments requires agents to perform semantic search under partial observability. 
Beyond recognizing the target object, an agent must infer promising search directions from incomplete egocentric observations, spatial layouts, and commonsense regularities of human environments. 
This makes high-level navigation decisions inherently semantic: a robot looking for a microwave should prioritize regions that visually and geometrically resemble kitchens, while one searching for a lamp should reason about bedrooms, desks, and living areas. 
Large language models (LLMs) and vision-language models (VLMs)~\cite{achiam2023gpt,team2023gemini,anthropic2024claude} offer a natural source of such semantic-spatial priors, motivating recent zero-shot navigation methods that leverage foundation models without environment-specific training~\citep{cai2024bridging, zhang2025apexnav, gong2026stairway}.

However, how to interface foundation models with a navigation system remains unresolved. 
One line of work directly queries LLMs or VLMs during navigation to predict actions~\citep{nie2025wmnav}, values~\citep{yokoyama2024vlfm}, or exploration directions from raw observations. 
While this preserves rich visual evidence for semantic reasoning, it requires frequent model inference; using strong VLMs such as GPT-4o~\citep{hurst2024gpt} becomes prohibitively slow for real robots, whereas lightweight vision-language models~\cite{li2023blip} often lack the semantic and spatial reasoning ability needed for complex indoor environments. 
Another line of work reduces inference cost by first abstracting the environment into textual summaries, semantic maps, topological memories, or scene graphs, and then reasoning over these compact structures~\cite{raychaudhuri2025semantic}. 
Although such abstractions are effective for memory and planning, they decouple high-level decisions from the raw visual evidence that originally supports them, making the navigation policy heavily dependent on the accuracy and completeness of upstream abstraction modules~\citep{yang20253d}.

This suggests that the bottleneck of VLM-based navigation lies not only in the capability of foundation models themselves, but also in the decision interface through which their semantic-spatial priors are exposed to robot navigation. 
We argue that the key challenge is not simply whether to use VLMs for navigation, but when and how they should be invoked. 
VLMs should not act as dense low-level controllers, nor should they be restricted to reasoning over prematurely compressed symbolic memories~\citep{yin2025unigoal}. 
Instead, an effective interface should satisfy three requirements: it should invoke VLMs only at decision-critical moments, expose raw visual observations that preserve semantic cues, and constrain the model's output to geometrically executable navigation goals. 
Such an interface would allow strong VLMs to perform deep semantic-spatial reasoning where it matters, while leaving continuous motion execution to efficient low-level planners.

\begin{figure}[t]
\centering
\includegraphics[width=0.9\columnwidth]{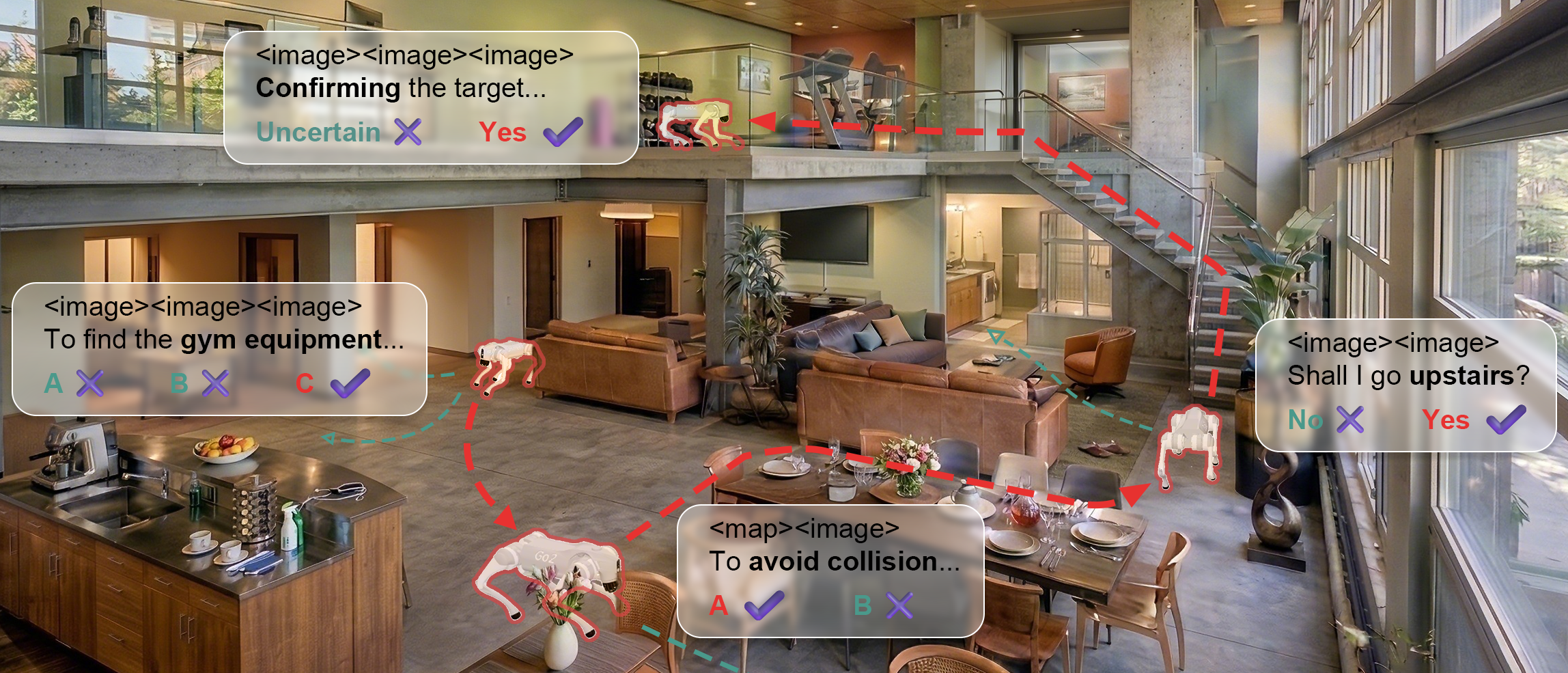} 
\caption{
PIGEON uses Points of Interest as sparse visual decision units that couple executable waypoints with raw observations, allowing VLMs to reason only at critical navigation moments.
}
\label{teaser}
\vspace{-1.5em}
\end{figure}

To this end, we propose PIGEON, a VLM-driven object navigation framework built around Points of Interest (PoIs). 
In object navigation, each high-level decision must jointly specify where the robot can move, what visual evidence supports this move, and why this move is useful for finding the target. 
PoIs are designed as the minimal decision units that bind these factors together. 
A PoI is therefore not merely a waypoint, but a sparse visual decision unit that couples a geometrically executable navigation goal with raw egocentric observations collected around it. 
PIGEON instantiates PoIs for exploration frontiers, suspected target objects, traversable stairs, and floor-level summaries, thereby unifying local exploration, target verification, and floor transition under a single sparse selection formulation. 
At each high-level decision step, the VLM compares candidate PoIs using their associated raw observations and selects the next waypoint, while a low-level planner executes continuous motion between PoIs. 
This design preserves the visual evidence needed for semantic-spatial reasoning, yet avoids dense model inference by invoking the VLM only at critical decision points.

Beyond efficient inference, the PoI interface also makes high-level navigation decisions verifiable. 
Because each candidate PoI is grounded in the map and associated with visual evidence, its utility can be automatically evaluated using signals such as shortest-path distance to the target or target visibility. 
We therefore introduce a Reinforcement Learning with Verifiable Rewards (RLVR)~\citep{wen2026reinforcement} pipeline tailored to sparse PoI selection, improving task-specific spatial reasoning in locally deployable VLMs without manually annotated Chain-of-Thought supervision. 
Extensive experiments on Habitat ObjectNav benchmarks show that PIGEON achieves state-of-the-art zero-shot performance, scales consistently with foundation model capacity, and transfers to Active Embodied Question Answering with only prompt modifications. 
We further validate the framework on physical robots, demonstrating that sparse raw-observation-grounded decision making improves both navigation robustness and practical deployment efficiency. 
Discussions of broader related studies are provided in Appendix \ref{app:related_work}.

Our contributions are summarized as follows:
\begin{itemize}[leftmargin=2.5em]
    \item We formulate VLM-based object navigation as raw-observation-grounded sparse decision making, where PoIs serve as geometrically executable and visually grounded decision units.
    \item We instantiate this in PIGEON, a zero-shot framework unifying frontier exploration, target verification, stair navigation, and floor reasoning via sparse PoI selection.
    \item We develop an automated RLVR fine-tuning pipeline for PoI selection, equipping smaller local VLMs with navigation-specific spatial reasoning without manual CoT annotations.
\end{itemize}



	


\section{Method}

\subsection{Problem Formulation}
In a navigation episode, the agent is initialized at a specific location on the map, tasked with navigating close to a certain object category within a given number of steps. At each timestep $t$, the agent receives an observation $O_t=(r_t, d_t, p_t)$, representing the RGB image, the depth image, and the agent's position, respectively. The position information $p_t$ can be provided by an external odometry or estimated by a localization module, e.g., a SLAM system. The agent must then output an action $a_t$, which can be either discrete or continuous, depending on the robot configuration.

To reduce the intervention frequency of VLM, we formulate the task as a waypoint navigation problem. The agent's primary objective is to traverse a waypoint sequence, denoted as $(\mathbf{p}_1, \mathbf{p}_2, \dots, \mathbf{p}_n)$. 
Upon reaching a waypoint $\mathbf{p}_k$, the agent reasons to determine the subsequent waypoint, $\mathbf{p}_{k+1}$. A low-level planner such as a PointNav RNN \cite{yokoyama2024vlfm} is engaged to facilitate navigation from $\mathbf{p}_k$ to $\mathbf{p}_{k+1}$. 
The process concludes when the agent reaches the final destination $\mathbf{p}_n$, where a stop command is issued.

\subsection{PoI Generation}
In our method, every waypoint in the sequence belongs to a PoI, thus the whole navigation process can be modeled as analyzing and navigating between PoIs. To facilitate the navigation by VLM in critical decision moments, we introduce our PoI categories as follows.

\textbf{Frontier PoI}\quad At each navigation step $t$, the agent maintains a BEV occupancy grid map and an exploration map to generate exploration frontiers like VLFM~\cite{yokoyama2024vlfm}. If an observation $O_t$ expands the explored area, a non-directional PoI on the boundary of the newly expanded area is created. 

\textbf{Suspect Object PoI}\quad Concurrently, we use an object detector and a segmentation model to identify and segment objects in the image that are similar to the navigation target. After the detection and segmentation, the suspect object 3D point cloud with a detection confidence will be generated, and fused with spatially overlapping objects, similar to ApexNav~\cite{zhang2025apexnav}. If a potential target object has a confidence exceeding a threshold $\tau_{sus}$ and is the highest among all associated objects categories, a Suspect Object PoI facing the object is created in a traversable area around the object. If the confidence exceeds $\tau_{target}$, the object is recognized as the target with its 2D position $\mathbf{p}_n$.

\textbf{Stair PoI and Floor Summary PoI}\quad During exploration, a height grid map is updated, and grids whose height is lower than $h_{floor}$ are recognized as a sinking region. To enable multi-layer exploration, the staircases are regarded as a special object category. After the confidence of a staircase exceeds $\tau_{stair}$, VLM examine the current RGB image and report the stair type if visible. If it leads upstairs or downstairs in a sinking region, the edge of stair region with several observation images will be integrated into a Stair PoI. Apart from Stair PoI, we also integrate the floor information within Summary PoI, which we will further discuss in \ref{section/multi-floor}.

In conclusion, we integrate frontiers, suspect objects, stairs, and floor summaries, into a uniform representation, PoI, which consists of a geometrical waypoint and several semantical observations.

\begin{figure}[t]
\centering
\includegraphics[width=1\columnwidth]{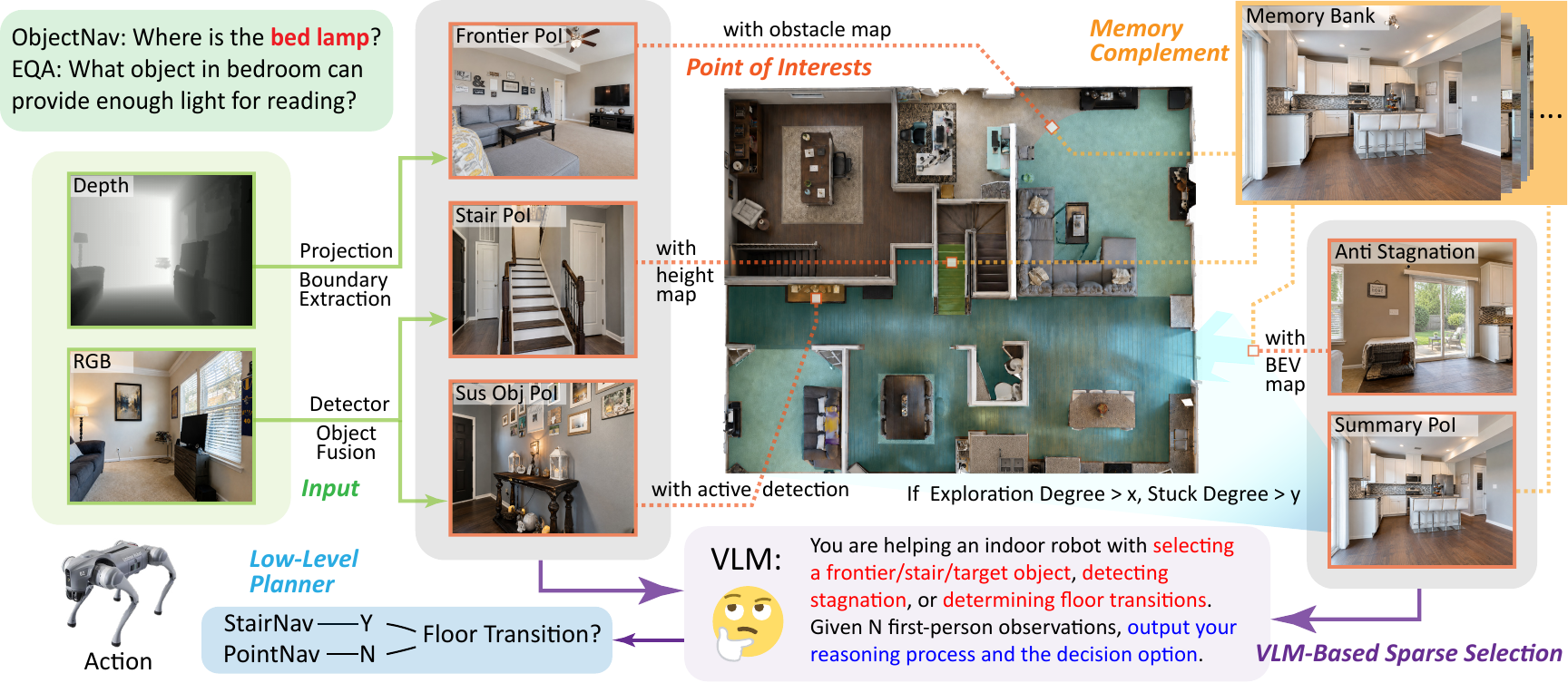} 
\caption{The overview of PIGEON. During exploration, several types of PoIs are constructed dynamically to enable fine-grained exploration and inter-floor navigation. Upon reaching a previously selected PoI, the VLM extracts informative PoIs from memory and selects the most relevant PoI as the next waypoint. A low-level planner is employed subsequently.}
\label{illustration}
\vspace{-1.0em}
\end{figure}

\subsection{PoI Management}

Although we introduced raw observation images in PoIs, upstream models including occupancy map and object detector are still needed for efficiency. Furthermore, raw observations which fail to serve as VLM input will be abandoned, leading to information loss. To address these issues, we introduce several ways to manage PoIs, enriching the robustness and semantic utilization of PIGEON.

\textbf{Memory Complement}\quad At each step of navigation, we record the capture position, angle, and BEV frustum of the RGB image $r_t$ as a memory to complement PoI observations. We formulate the memory complement into a Maximum Coverage Problem (MCP), which is solved via a greedy heuristic to select eligible observations that maximize the union coverage of their frustums~\citep{nemhauser1978analysis}.

\textbf{Anti Stagnation}\quad Once the steps navigating to the target PoI exceeds $t_{stuck}$, we retrieve the reference images around the waypoint using MCP, and feed the reference images into the VLM, with a colored BEV map containing the robot’s trajectory. If the VLM determines that further navigation to the POI is unnecessary, the PoI and its vicinity will be marked as impassable.

\textbf{VLM-based PoI Selection}\quad We adopt a method similar to Mem2Ego\cite{zhang2025mem2ego} to project PoI markers onto 2D observations, including Frontier PoIs and Stair PoIs. For each PoI $p_i$, we project the PoI's map location into the corresponding image with circular numeric markers, and feed those images into the VLM for selection. If the VLM determines that the current information is insufficient to make a decision, it outputs a special token (e.g., 0) as instructed.

\subsection{Multi-Floor Policy}
\label{section/multi-floor}
\begin{figure}[t]
\centering
\includegraphics[width=1\columnwidth]{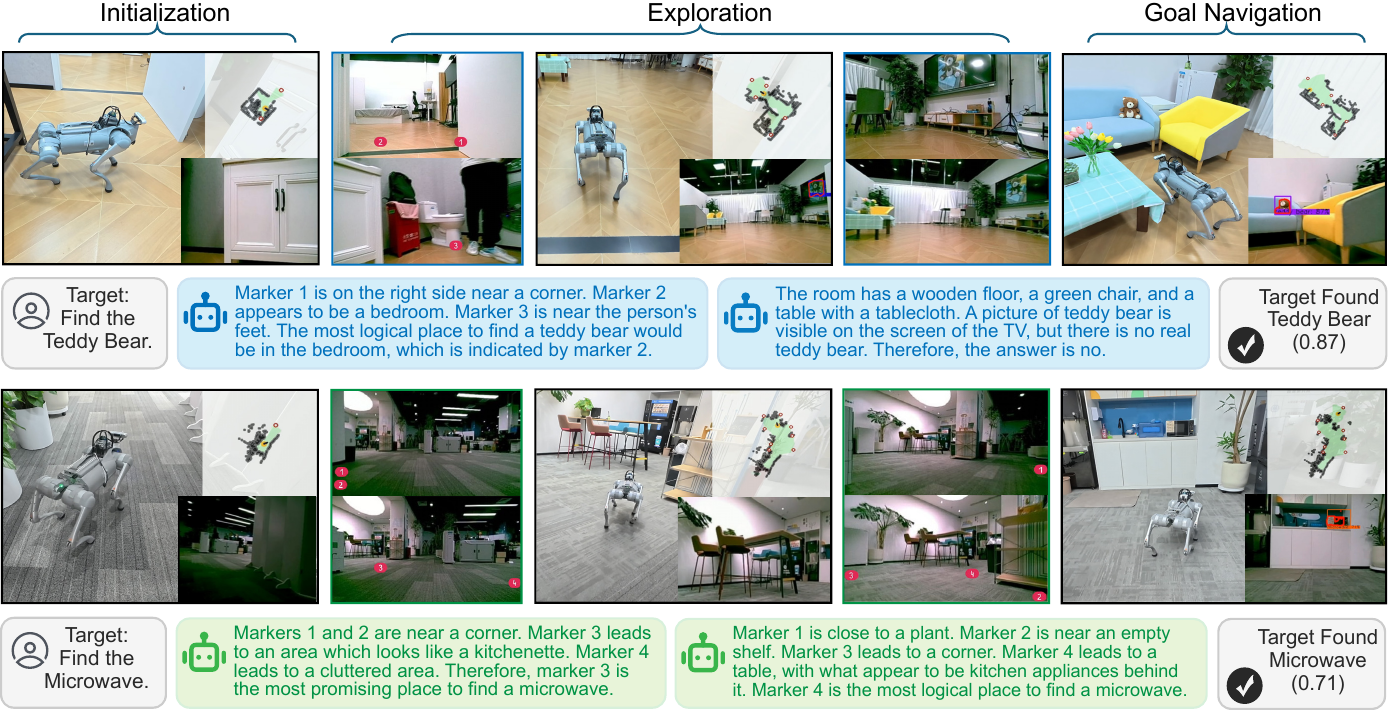} 
\caption{Real-world deployment of PIGEON on a physical Unitree Go2 quadruped across diverse environments. See Appendix \ref{app:real_world} for detailed experimental setups and additional hardware evaluations.}
\label{illustration}
\vspace{-1.0em}
\end{figure}



Current navigation methods typically navigate only within a single floor. While ASCENT~\citep{gong2026stairway} pioneers zero-shot multi-floor object navigation, its reliance on fixed step-count thresholds for floor switching and textual summarizations for LLM inference risk the loss of environment details.

Given current exploration step $t$, exploration area $S$, boundary length $L$, we define two metrics: \textbf{Exploration Degree} $S / L$ and \textbf{Stuck Degree} $t / S$. An increase in the Exploration Degree indicates a larger explored area with fewer frontiers, whereas an increase in the Stuck Degree reflects prolonged lingering within a confined region. If the Exploration Degree exceeds $\alpha_{exp}$, or Stuck Degree exceeds $\alpha_{stuck}$, $n_{floor}$ observations on the current floor will be gathered using MCP algorithm. These observations serve as floor Summary PoI, whose geometric waypoint is located on the floor above. After the inference of VLM on floor summaries, the output signal of VLM triggers the floor transition procedure, including navigating to the selected Stair PoI, and climbing up or down the stairs.

\subsection{RLVR Pipeline}

Our method can be fine-tuned via reinforcement learning on datasets providing 2D BEV maps and object annotations, circumventing the need for expensive manual Chain-of-Thought (CoT) data. We formulate training samples as tuples $(v, \mathcal{T}, y)$, where $v$ is the visual observation, $\mathcal{T}$ is the text instruction, and $y$ the verifiable ground truth. For Frontier PoI selection, $y = \mathbf{D}$, defined as a list of shortest distances $d_j$ from each candidate PoI $j$ to the target. For object confirmation, $y$ is the target's visible pixel count.

The VLM is prompted to generate reasoning steps \texttt{<think>...\allowbreak</think>} before yielding a final decision \texttt{<action>...\allowbreak</action>}. To enrich the dataset with sub-optimal recovery behaviors, we collect trajectories using an $\epsilon$-greedy policy, selecting the optimal PoI (minimum $d_j$) with probability $1-\epsilon$ and a random candidate otherwise. 

We employ Group Relative Policy Optimization (GRPO)~\cite{shao2024deepseekmath} to optimize the policy. For a given input query, we sample a group of $G$ responses. When the task is Frontier PoI selection, we extract the predicted PoI index $j$ from the $i$-th response and compute a soft, distance-scaled reward $R_i$:
\begin{gather}
A_i=\frac{R_i-\mu\{R_1,\dots, R_G\}}{\sigma\{R_1,\dots,R_G\}},\quad R_i=
r_j=\begin{cases}
(d_{\max}-d_j)/(d_{\max}-d_{\min})& 1\le j \le n\\
\mathrm{Avg}_{1\le k\le n}[r_k] & \text{otherwise}
\label{reward_design}
\end{cases}
\end{gather}
Here, $d_{\max}$ and $d_{\min}$ represent the extreme distances in $\mathbf{D}$. The \textit{otherwise} case handles uncertain outputs by assigning the expected reward of a random valid guess. Conversely, for object confirmation tasks, $R_i$ is assigned a binary reward with uncertain fallback based on pixel count matching. 
\section{Experiments}
\subsection{Experimental Setting}
We evaluate our method in the Habitat simulator \cite{savva2019habitat}, using datasets from the 2021-2023 Habitat Challenges: MP3D \cite{Matterport3D}, HM3Dv1 \cite{ramakrishnan2021habitat}, and HM3Dv2 \cite{yadav2023habitat}. The MP3D validation set contains 2195 episodes across 11 scenes, including a 512-episode subset (MP3D Mul) requiring multi-floor navigation. The HM3Dv1 validation set includes 2000 episodes across 20 scenes with a 429-episode multi-floor subset (HM3D Mul). The HM3Dv2 validation set consists of 1000 episodes across 10 scenes, where targets remain on the same floor. We use YOLOv7 \cite{wang2023yolov7} for object detection in COCO classes and OWLv2 \cite{minderer2023scaling} for other object categories. For all experiments, we use Mobile SAM \cite{mobile_sam} as the segmentation model.
For all compared methods, we report the Success Rate (SR), and Success weighted by inverse Path Length (SPL) which marks the navigation efficiency.

At the beginning of each episode, the agent performs a full rotation to gather initial observations. During the intermediate steps of navigating from a waypoint $\mathbf{p}_k$ to $\mathbf{p}_{k+1}$, we rely exclusively on a low-level planner like PointNav for action control. For floor switching, we trained a StairNav RNN for action execution using Variable Experience Rollout (VER)~\citep{wijmans2022ver} as detailed in Appendix~\ref{app:rl}. In terms of high-level decision-making VLMs, we use GPT-4o-2024-02-15-preview for our non-RL approach \textit{PIGEON-ZeroShot} and open-source Qwen-2.5VL-7B~\citep{bai2025qwen2} for RLVR evaluation.

\subsection{Main Results}
\begin{table}[b]
  \centering
  \footnotesize
  \caption{Comparison with SOTA methods which are categorized by their required inputs, and the form of intermediate representations passed to the main decision model. (No: The main decision model utilize raw observations. Struct. Text: Structured Text, Synth. Views: Synthetic Views)}
  \setlength{\tabcolsep}{2pt}
  \label{tab:performance}
  \begin{tabular}{llccccccccr}
    \toprule
    \multirow{2}{*}{\textbf{Method}} & \multirow{2}{*}{\textbf{Venue}} & \multirow{2}{*}{\makecell{\textbf{Training}\\\textbf{Free}}} & \multirow{2}{*}{\makecell{\textbf{Oracle}\\\textbf{Free}}} & \multicolumn{2}{c}{\textbf{HM3D v1}} & \multicolumn{2}{c}{\textbf{HM3D v2}} & \multicolumn{2}{c}{\textbf{MP3D}} & \multirow{2}{*}{\makecell[r]{\textbf{Intermediate}\\\textbf{Representation}}} \\
    \cmidrule(lr){5-6} \cmidrule(lr){7-8} \cmidrule(lr){9-10}
    & & & & SR$\uparrow$ & SPL$\uparrow$ & SR$\uparrow$ & SPL$\uparrow$ & SR$\uparrow$ & SPL$\uparrow$ & \\
    \midrule
    PONI~\cite{ramakrishnan2022poni} & CVPR'22 & No & Yes & - & - & - & - & 31.8 & 12.1 & \makecell[r]{Semantic BEV} \\
    PIRLNav~\cite{ramrakhya2023pirlnav} & CVPR'23 & No & Yes & 70.4 & 34.1 & - & - & - & - & Implicit \\
    Uni-Navid~\cite{zhang2024uni} & RSS'25 & No & Yes & \textbf{73.7} & \textbf{37.1} & - & - & - & - & Implicit \\
    ImagineNav++* & TPAMI'26 & Yes & No & 62.5 & 32.8 & - & - & - & - & No \\
    STRIVE~\cite{zhu2025strive} & ICRA'26 & Yes & No & 62.9 & 34.2 & \textbf{79.6} & \textbf{38.7} & \textbf{52.3} & \textbf{23.1} & \makecell[r]{Struct. Text} \\
    \midrule
    VLFM~\cite{yokoyama2024vlfm} & ICRA'24 & Yes & Yes & 52.5 & 30.4 & 63.6 & 32.5 & 36.4 & 17.5 & No \\
    InstructNav~\cite{long2024instructnav} & CoRL'24 & Yes & Yes & - & - & 58.0 & 20.9 & - & - & \makecell[r]{Struct. Text} \\
    ApexNav~\cite{zhang2025apexnav} & RAL'25 & Yes & Yes & 59.6 & 33.0 & \underline{76.2} & \underline{38.0} & 39.2 & \underline{17.8} & No \\
    UniGoal~\cite{yin2025unigoal} & CVPR'25 & Yes & Yes & 54.5 & 25.1 & - & - & 41.0 & 16.4 & \makecell[r]{Struct. Text} \\
    WMNav~\cite{nie2025wmnav} & \makecell[l]{IROS'25 Oral} & Yes & Yes & 58.1 & 31.2 & 72.2 & 33.3 & \underline{45.4} & 17.2 & No \\
    ASCENT~\cite{gong2026stairway} & RAL'26 & Yes & Yes & \underline{65.4} & \underline{33.5} & - & - & 44.5 & 15.5 & \makecell[r]{Struct. Text} \\
    ImagineNav++~\cite{wang2026imaginenav++} & TPAMI'26 & Yes & Yes & 58.5 & 26.6 & - & - & - & - & \makecell[r]{Synth. Views} \\
    \textbf{\textit{PIGEON-ZeroShot}} & \textbf{-} & \textbf{Yes} & \textbf{Yes} & \textbf{67.1} & \textbf{35.8} & \textbf{79.6} & \textbf{38.1} & \textbf{52.6} & \textbf{18.5} & \textbf{No} \\
    \bottomrule
  \end{tabular}
\end{table}

As shown in Table \ref{tab:performance}, compared with other methods, our method surpasses previous zero-shot models on most metrics. Meanwhile, on HM3D v2 and MP3D datasets, the performance of our method is very close to latest SoTA STRIVE which uses oracle path planner, except SPL on MP3D.

\textbf{Methods with raw representations generally perform better than other training-free methods with compressed representations.} Apart from \textit{PIGEON-ZeroShot}, intermediate-representation-free method ApexNav ranks the second best performance on HM3D v2 dataset, and WMNav ranks the second best on MP3D dataset. On HM3D v1, ApexNav still performs competitively, while ASCENT which leverages text as LLM input ranks the second best. However, we attribute it to the multi-level strategy of ASCENT for HM3D Mul subset. 
ImagineNav++ also achieved a significant performance improvement after the representations were replaced with oracle observations, as indicated by *.

\textbf{Methods leveraging larger models like LLMs or VLMs generally perform better than other methods.} In training-free methods, dense value map are both employed by VLFM and ApexNav, while ApexNav utilizes DeepSeek-V3 for similar object inference, leading to better performance results. Our approach further improved performance through the enhanced use of VLMs. However, due to less VLM intervention frequency, the gains in SPL are not as pronounced as the gains in SR. Among training-based methods, Uni-Navid also outperforms PIRLNav which has fewer parameters.

\textbf{The model's performance varies depending on the navigable area size and annotation quality in the dataset.} On HM3D v2, we achieve 3.4\% SR improvement over ApexNav. We also achieve 1.7\% SR and 2.3\% SPL improvements on HM3D v1, and 7.2\% SR improvements on MP3D. HM3D v1 and MP3D include multi-floor navigation episodes, while MP3D features larger navigable areas but lower annotation quality compared to HM3D. We suggest that more complex environments enable our approach to leverage the relative advantages of deep reasoning over other methods.

\subsection{RLVR Results}

\begin{table}[h]
\caption{Evaluation Metrics of progressively enhanced RLVR configuration on HM3D v2 dataset.} 
\small
\label{tab:rlvr_results}
\centering
\begin{tabular}{lcccccc}
    \toprule
    \textbf{Configuration} & \textbf{SR}$\uparrow$ & \textbf{SPL}$\uparrow$ & \textbf{Entropy}$\uparrow$ & \textbf{Avg Len}$\uparrow$ & \textbf{\# Out=0}$\downarrow$ & \textbf{\# Unconf.}$\downarrow$ \\ 
    \midrule
    Vanilla 7B & 72.66 & 33.16 & 0.5205 & 53.67 & 92 & 94 \\
    \midrule
    \textbf{w} Format Reward & 73.00 & 33.71 & 0.8466 & \textbf{94.85} & 182 & 93 \\
    \textbf{w} All Reward & 73.50 & 35.02 & 0.9075 & 78.02 & 5 & \underline{2} \\
    \textbf{w} Online Filtering & 74.00 & 34.99 & 0.9615 & 76.14 & 10 & 3 \\
    \textbf{w} Long Training & 75.72 & \textbf{36.99} & 0.7779 & 81.28 & 3 & 58 \\
    \textbf{w} 30K Data & \underline{76.90} & \underline{36.56} & \underline{0.9983} & 81.58 & \textbf{0} & 34 \\
    \textbf{w/o} KL Loss & \textbf{77.56} & 36.55 & \textbf{1.4393} & \underline{86.85} & \textbf{0} & \textbf{0} \\
    \bottomrule
\end{tabular}
\end{table}

As shown in Table \ref{tab:rlvr_results}, the navigation performance has steadily improved with the introduction of new training elements, including training rewards, training scales, and techniques such as online filtering (i.e., dynamic sampling in DAPO~\citep{yu2026dapo}) and removing KL loss~\citep{guo2025deepseek}. While some studies~\citep{yue2025does} argue that RLVR merely boosts model confidence (manifested as decreased entropy) without imparting new capabilities, our generation-specific statistics tell a different story. We report the average token entropy, average output length, and metrics for uncertain Frontier PoI selections (\texttt{\#Out=0}) and unconfirmed Object PoIs (\texttt{\#Unconf.}).

\textbf{RLVR training on sparse PoI selection enhances the model's capabilities rather than merely boosting its confidence.} As the training configuration expands, the count of output uncertainty decreases, indicating that the model has greater confidence in its answers. However, the model’s entropy and average length show an overall upward trend, demonstrating that this improvement does not come at the expense of mode collapse or overconfidence.

\subsection{Ablation Study} 


\begin{table}[b]
\centering
\small
\caption{Ablation study of our method's key components on the HM3D and MP3D dataset. Mul indicates multi-floor scenarios. Datasets difficulty increase progressively from left to right.}
\label{tab:combined_ablation}
\setlength{\tabcolsep}{3pt} 
\begin{tabular}{@{} l cccc l cccc @{}}
    \toprule 
    \textbf{Method} & \multicolumn{2}{c}{\textbf{HM3D v2}} & \multicolumn{2}{c}{\textbf{HM3D v1}} & \textbf{Method} & \multicolumn{2}{c}{\textbf{HM3D Mul}} & \multicolumn{2}{c}{\textbf{MP3D Mul}} \\
    \midrule
    \textit{PIGEON-ZeroShot} & 79.6 & 38.1 & 67.1 & 35.8 & - & 35.9 & 16.1 & 38.5 & 14.1 \\
    \midrule
    w/o Frontier PoI    & 78.3 & 37.1 & 66.5 & 34.6 & - & 34.5 & 14.2 & 34.2 & 11.3  \\
    w/o Object PoI      & 69.2 & 32.8 & 56.2 & 28.5 & w/o Stair Summary & 25.2 & 9.1  & 20.3 & 8.8  \\
    w/o Mem. Complement & 75.3 & 38.0 & 63.7 & 32.1 & w/o Stair Frontier & 31.9 & 13.6 & 33.8 & 12.8 \\
    w/o Multi-Floor     & 77.1 & 37.5 & 60.6 & 32.3 & w/o Stag. Detection & 34.7 & 15.2 & 29.5 & 11.7 \\
    \bottomrule 
    \end{tabular}
\end{table}

As shown in Table \ref{tab:combined_ablation}, removing Frontier PoI or Stagnation Detection causes steeper performance drops on the complex MP3D dataset than on HM3D v2. This indicates that larger search areas rely more heavily on the VLM's semantic priors, whereas limited areas can be largely solved by geometric exploration. Conversely, removing Object PoI uniformly degrades performance on HM3D v1 and v2 datasets, confirming its fundamental role in target verification regardless of environment size.

The HM3D v2 episodes consist only single floor, thus removing the Multi-Floor module has only a limited effect on its performance. Similarly, after removing the Memory Complement module, only the success rate on HM3D v2 showed a significant decline. For Stair Summary and Stair Frontier which are influenced by the Memory Complement module, we conducted separate tests on HM3D Mul and MP3D Mul, where the performance declined significantly following ablation.

\subsection{Further Analysis}


In this section, we provide a deeper analysis of our proposed framework from two key perspectives, i.e., the scalability and the zero-shot transferability of our architecture.

\begin{figure}[t] 
    \centering       
    \includegraphics[width=\linewidth]{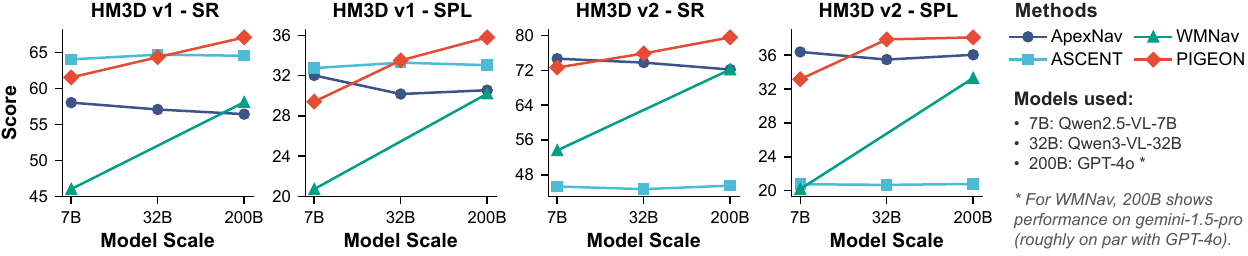} 
    \caption{Scaling capability comparison in HM3D dataset.} 
    \label{fig:scaling}
    \vspace{-1.0em}
\end{figure}

\textbf{Methods with raw representations exhibit strong scalability with foundation model capabilities.} As illustrated in Figure~\ref{fig:scaling}, the performance metrics of PIGEON and WMNav, which utilize raw representations, consistently improve as the underlying model scales up. Unlike approaches that rely heavily on task-specific abstractions, our raw-representation paradigm avoids information bottlenecks.

\textbf{Our method can be transferred to other navigation tasks with simple modifications to the prompt.} To validate transferability, we evaluate PIGEON on the Active Embodied Question Answering (A-EQA) task under the OpenEQA-184 setting~\citep{majumdar2024openeqa, yang20253d}.

\begin{wraptable}{r}{0.40\linewidth}
    \vspace{-20pt}
    \setlength{\tabcolsep}{2pt} 
    \centering
    \caption{Model comparison in A-EQA} \label{tab:aeqa_performance}
    \resizebox{\linewidth}{!}{
        \begin{tabular}{lcc}
            \toprule
            Method & LLM-SR$\uparrow$ & LLM-SPL$\uparrow$ \\
            \midrule
            FAST-EQA~\citep{zhang2026fast} & 49.0  & 27.7 \\
            MTU3D~\cite{zhu2025mtu} & 51.1  & 42.6 \\
            3D-Mem~\cite{yang20253d} & 52.6  & 42.0 \\
            \textit{PIGEON-ZeroShot} & 52.5  & 47.4 \\
            \bottomrule
        \end{tabular}%
    }
    \vspace{-10pt}
\end{wraptable}

As shown in Table~\ref{tab:aeqa_performance}, PIGEON achieves competitive LLM-SR and significantly superior SPL without any task-specific fine-tuning. This high SPL demonstrates that our prompt-driven sparse waypoint selection generates highly efficient trajectories for information gathering, highlighting its potential as a generalist embodied architecture.




\section{Limitations and Future Work}

While our proposed framework demonstrates strong performance and generalization capabilities, it has several limitations that present promising avenues for future research.

\textbf{Detector Reliance}\quad Relying on open-vocabulary models (e.g., OWLv2) occasionally causes localization ambiguities for structural categories like stairs, increasing VLM computational overhead. 

\textbf{Single-Target Focus}\quad Extending our formulation to multi-target or sequential scenarios requires persistent spatial memory with raw observations. Future work could integrate episodic memory architectures~\citep{yang20253d, anwar2025remembr} for long-horizon visual retrieval. 

\textbf{Representation Trade-offs}\quad While raw observations successfully avoid information bottlenecks, for grounding complex language landmarks or strictly minimizing reasoning overhead, intermediate representations remain highly efficient. Developing a hybrid paradigm that adaptively fuses lossless context with intermediate states remains a critical direction to further elevate embodied agents.
\section{Conclusion}
In this paper, we introduced PIGEON, a novel object navigation framework that redefines the VLM-Robot decision interface to resolve the dilemma between computational efficiency and high-level spatial reasoning. By formulating navigation as a sparse PoI selection problem directly using raw visual observations, this interface effectively bypasses the information bottlenecks inherent in abstract representations. Extensive evaluations demonstrate that PIGEON achieves state-of-the-art performance, seamlessly pairs with an RLVR fine-tuning pipeline, scales consistently with foundation models, and generalizes zero-shot to both A-EQA tasks and distinct real-world robotic platforms.



\bibliography{corl2026}  

@article{yu2026dapo,
  title={Dapo: An open-source llm reinforcement learning system at scale},
  author={Yu, Qiying and Zhang, Zheng and Zhu, Ruofei and Yuan, Yufeng and Zuo, Xiaochen and Yue, Yu and Dai, Weinan and Fan, Tiantian and Liu, Gaohong and Liu, Lingjun and others},
  journal={Advances in Neural Information Processing Systems},
  volume={38},
  pages={113222--113244},
  year={2026}
}

@inproceedings{
    wen2026reinforcement,
    title={Reinforcement Learning with Verifiable Rewards Implicitly Incentivizes Correct Reasoning in Base {LLM}s},
    author={Xumeng Wen and Zihan Liu and Shun Zheng and Shengyu Ye and Zhirong Wu and Yang Wang and Zhijian Xu and Xiao Liang and Junjie Li and Ziming Miao and Jiang Bian and Mao Yang},
    booktitle={The Fourteenth International Conference on Learning Representations},
    year={2026},
    url={https://openreview.net/forum?id=jGbRWwIidy}
}

@article{guo2025deepseek,
  title={DeepSeek-R1 incentivizes reasoning in LLMs through reinforcement learning},
  author={Guo, Daya and Yang, Dejian and Zhang, Haowei and Song, Junxiao and Wang, Peiyi and Zhu, Qihao and Xu, Runxin and Zhang, Ruoyu and Ma, Shirong and Bi, Xiao and others},
  journal={Nature},
  volume={645},
  number={8081},
  pages={633--638},
  year={2025},
  publisher={Nature Publishing Group UK London}
}

@inproceedings{majumdar2024openeqa,
  title={Openeqa: Embodied question answering in the era of foundation models},
  author={Majumdar, Arjun and Ajay, Anurag and Zhang, Xiaohan and Putta, Pranav and Yenamandra, Sriram and Henaff, Mikael and Silwal, Sneha and Mcvay, Paul and Maksymets, Oleksandr and Arnaud, Sergio and others},
  booktitle={Proceedings of the IEEE/CVF conference on computer vision and pattern recognition},
  pages={16488--16498},
  year={2024}
}

@inproceedings{zhang2026fast,
  title={FAST-EQA: Efficient Embodied Question Answering with Global and Local Region Relevancy},
  author={Zhang, Haochen and Savaliya, Nirav and Siddiqui, Faizan and Sachdeva, Enna},
  booktitle={Proceedings of the IEEE/CVF Winter Conference on Applications of Computer Vision},
  pages={1664--1673},
  year={2026}
}

@inproceedings{li2023blip,
  title={Blip-2: Bootstrapping language-image pre-training with frozen image encoders and large language models},
  author={Li, Junnan and Li, Dongxu and Savarese, Silvio and Hoi, Steven},
  booktitle={International conference on machine learning},
  pages={19730--19742},
  year={2023},
  organization={PMLR}
}

@inproceedings{
    wu2024voronav,
    title={VoroNav: Voronoi-based Zero-shot Object Navigation with Large Language Model},
    author={Pengying Wu and Yao Mu and Bingxian Wu and Yi Hou and Ji Ma and Shanghang Zhang and Chang Liu},
    booktitle={Forty-first International Conference on Machine Learning},
    year={2024},
    url={https://openreview.net/forum?id=Va7mhTVy5s}
}

@inproceedings{wang2023dreamwalker,
  title={Dreamwalker: Mental planning for continuous vision-language navigation},
  author={Wang, Hanqing and Liang, Wei and Van Gool, Luc and Wang, Wenguan},
  booktitle={Proceedings of the IEEE/CVF international conference on computer vision},
  pages={10873--10883},
  year={2023}
}

@inproceedings{cai2024bridging,
  title={Bridging zero-shot object navigation and foundation models through pixel-guided navigation skill},
  author={Cai, Wenzhe and Huang, Siyuan and Cheng, Guangran and Long, Yuxing and Gao, Peng and Sun, Changyin and Dong, Hao},
  booktitle={2024 IEEE International Conference on Robotics and Automation (ICRA)},
  pages={5228--5234},
  year={2024},
  organization={IEEE}
}

@inproceedings{yu2023l3mvn,
  title={L3mvn: Leveraging large language models for visual target navigation},
  author={Yu, Bangguo and Kasaei, Hamidreza and Cao, Ming},
  booktitle={2023 IEEE/RSJ International Conference on Intelligent Robots and Systems (IROS)},
  pages={3554--3560},
  year={2023},
  organization={IEEE}
}

@inproceedings{shah2023navigation,
  title={Navigation with large language models: Semantic guesswork as a heuristic for planning},
  author={Shah, Dhruv and Equi, Michael Robert and Osi{\'n}ski, B{\l}a{\.z}ej and Xia, Fei and Ichter, Brian and Levine, Sergey},
  booktitle={Conference on Robot Learning},
  pages={2683--2699},
  year={2023},
  organization={PMLR}
}

@inproceedings{kuang2024openfmnav,
  title={Openfmnav: Towards open-set zero-shot object navigation via vision-language foundation models},
  author={Kuang, Yuxuan and Lin, Hai and Jiang, Meng},
  booktitle={Findings of the Association for Computational Linguistics: NAACL 2024},
  pages={338--351},
  year={2024}
}

@inproceedings{
    li2026questa,
    title={QuestA: Expanding Reasoning Capacity in {LLM}s via Question Augmentation},
    author={Jiazheng Li and Hongzhou Lin and Hong Lu and Kaiyue Wen and Zaiwen Yang and Jiaxuan Gao and Yi Wu and Jingzhao Zhang},
    booktitle={The Fourteenth International Conference on Learning Representations},
    year={2026},
    url={https://openreview.net/forum?id=3MifB0f7qR}
}

@article{huang2024gamap,
  title={Gamap: Zero-shot object goal navigation with multi-scale geometric-affordance guidance},
  author={Huang, Hao and Hao, Yu and Wen, Congcong and Tzes, Anthony and Fang, Yi and others},
  journal={Advances in Neural Information Processing Systems},
  volume={37},
  pages={39386--39408},
  year={2024}
}

@inproceedings{zhou2023esc,
  title={Esc: Exploration with soft commonsense constraints for zero-shot object navigation},
  author={Zhou, Kaiwen and Zheng, Kaizhi and Pryor, Connor and Shen, Yilin and Jin, Hongxia and Getoor, Lise and Wang, Xin Eric},
  booktitle={International Conference on Machine Learning},
  pages={42829--42842},
  year={2023},
  organization={PMLR}
}

@inproceedings{zhou2024navgpt,
  title={Navgpt-2: Unleashing navigational reasoning capability for large vision-language models},
  author={Zhou, Gengze and Hong, Yicong and Wang, Zun and Wang, Xin Eric and Wu, Qi},
  booktitle={European Conference on Computer Vision},
  pages={260--278},
  year={2024},
  organization={Springer}
}

@inproceedings{nie2025wmnav,
  title={Wmnav: Integrating vision-language models into world models for object goal navigation},
  author={Nie, Dujun and Guo, Xianda and Duan, Yiqun and Zhang, Ruijun and Chen, Long},
  booktitle={2025 IEEE/RSJ International Conference on Intelligent Robots and Systems (IROS)},
  pages={2392--2399},
  year={2025},
  organization={IEEE}
}

@article{chang2023goat,
  title={Goat: Go to any thing},
  author={Chang, Matthew and Gervet, Theophile and Khanna, Mukul and Yenamandra, Sriram and Shah, Dhruv and Min, So Yeon and Shah, Kavit and Paxton, Chris and Gupta, Saurabh and Batra, Dhruv and others},
  journal={arXiv preprint arXiv:2311.06430},
  year={2023}
}

@article{wang2026vlingnav,
  title={VLingNav: Embodied Navigation with Adaptive Reasoning and Visual-Assisted Linguistic Memory},
  author={Wang, Shaoan and Luo, Yuanfei and Chen, Xingyu and Luo, Aocheng and Li, Dongyue and Liu, Chang and Chen, Sheng and Zhang, Yangang and Yu, Junzhi},
  journal={arXiv preprint arXiv:2601.08665},
  year={2026}
}

@inproceedings{yokoyama2024vlfm,
  title={Vlfm: Vision-language frontier maps for zero-shot semantic navigation},
  author={Yokoyama, Naoki and Ha, Sehoon and Batra, Dhruv and Wang, Jiuguang and Bucher, Bernadette},
  booktitle={2024 IEEE International Conference on Robotics and Automation (ICRA)},
  pages={42--48},
  year={2024},
  organization={IEEE}
}

@inproceedings{
    long2024instructnav,
    title={InstructNav: Zero-shot System for Generic Instruction Navigation in Unexplored Environment},
    author={Yuxing Long and Wenzhe Cai and Hongcheng Wang and Guanqi Zhan and Hao Dong},
    booktitle={8th Annual Conference on Robot Learning},
    year={2024},
    url={https://openreview.net/forum?id=fCDOfpTCzZ}
}

@inproceedings{yin2025unigoal,
  title={Unigoal: Towards universal zero-shot goal-oriented navigation},
  author={Yin, Hang and Xu, Xiuwei and Zhao, Linqing and Wang, Ziwei and Zhou, Jie and Lu, Jiwen},
  booktitle={Proceedings of the Computer Vision and Pattern Recognition Conference},
  pages={19057--19066},
  year={2025}
}

@article{wijmans2022ver,
  title={Ver: Scaling on-policy rl leads to the emergence of navigation in embodied rearrangement},
  author={Wijmans, Erik and Essa, Irfan and Batra, Dhruv},
  journal={Advances in Neural Information Processing Systems},
  volume={35},
  pages={7727--7740},
  year={2022}
}

@article{zhang2025mem2ego,
  title={Mem2ego: Empowering vision-language models with global-to-ego memory for long-horizon embodied navigation},
  author={Zhang, Lingfeng and Liu, Yuecheng and Zhang, Zhanguang and Aghaei, Matin and Hu, Yaochen and Gu, Hongjian and Alomrani, Mohammad Ali and Bravo, David Gamaliel Arcos and Karimi, Raika and Hamidizadeh, Atia and others},
  journal={arXiv preprint arXiv:2502.14254},
  year={2025}
}

@article{lei2025gaussnav,
  title={Gaussnav: Gaussian splatting for visual navigation},
  author={Lei, Xiaohan and Wang, Min and Zhou, Wengang and Li, Houqiang},
  journal={IEEE Transactions on Pattern Analysis and Machine Intelligence},
  volume={47},
  number={5},
  pages={4108--4121},
  year={2025},
  publisher={IEEE}
}

@inproceedings{zhang2025mapnav,
  title={Mapnav: A novel memory representation via annotated semantic maps for vlm-based vision-and-language navigation},
  author={Zhang, Lingfeng and Hao, Xiaoshuai and Xu, Qinwen and Zhang, Qiang and Zhang, Xinyao and Wang, Pengwei and Zhang, Jing and Wang, Zhongyuan and Zhang, Shanghang and Xu, Renjing},
  booktitle={Proceedings of the 63rd Annual Meeting of the Association for Computational Linguistics (Volume 1: Long Papers)},
  pages={13032--13056},
  year={2025}
}

@inproceedings{yang20253d,
  title={3D-mem: 3D scene memory for embodied exploration and reasoning},
  author={Yang, Yuncong and Yang, Han and Zhou, Jiachen and Chen, Peihao and Zhang, Hongxin and Du, Yilun and Gan, Chuang},
  booktitle={Proceedings of the Computer Vision and Pattern Recognition Conference},
  pages={17294--17303},
  year={2025}
}

@article{zheng20263dgsnav,
  title={3DGSNav: Enhancing Vision-Language Model Reasoning for Object Navigation via Active 3D Gaussian Splatting},
  author={Zheng, Wancai and Chen, Hao and Lu, Xianlong and Ou, Linlin and Yu, Xinyi},
  journal={arXiv preprint arXiv:2602.12159},
  year={2026}
}

@inproceedings{busch2025one,
  title={One map to find them all: Real-time open-vocabulary mapping for zero-shot multi-object navigation},
  author={Busch, Finn Lukas and Homberger, Timon and Ortega-Peimbert, Jes{\'u}s and Yang, Quantao and Andersson, Olov},
  booktitle={2025 IEEE International Conference on Robotics and Automation (ICRA)},
  pages={14835--14842},
  year={2025},
  organization={IEEE}
}

@article{zhang2025apexnav,
  title={Apexnav: An adaptive exploration strategy for zero-shot object navigation with target-centric semantic fusion},
  author={Zhang, Mingjie and Du, Yuheng and Wu, Chengkai and Zhou, Jinni and Qi, Zhenchao and Ma, Jun and Zhou, Boyu},
  journal={IEEE Robotics and Automation Letters},
  year={2025},
  publisher={IEEE}
}

@article{qi2025vln,
  title={VLN-R1: Vision-Language Navigation via Reinforcement Fine-Tuning},
  author={Qi, Zhangyang and Zhang, Zhixiong and Yu, Yizhou and Wang, Jiaqi and Zhao, Hengshuang},
  journal={arXiv preprint arXiv:2506.17221},
  year={2025}
}

@article{feige1998threshold,
  title={A threshold of ln n for approximating set cover},
  author={Feige, Uriel},
  journal={Journal of the ACM (JACM)},
  volume={45},
  number={4},
  pages={634--652},
  year={1998},
  publisher={ACM New York, NY, USA}
}

@article{nemhauser1978analysis,
  title={An analysis of approximations for maximizing submodular set functions—I},
  author={Nemhauser, George L and Wolsey, Laurence A and Fisher, Marshall L},
  journal={Mathematical programming},
  volume={14},
  number={1},
  pages={265--294},
  year={1978},
  publisher={Springer}
}

@inproceedings{gu2024conceptgraphs,
  title={Conceptgraphs: Open-vocabulary 3d scene graphs for perception and planning},
  author={Gu, Qiao and Kuwajerwala, Ali and Morin, Sacha and Jatavallabhula, Krishna Murthy and Sen, Bipasha and Agarwal, Aditya and Rivera, Corban and Paul, William and Ellis, Kirsty and Chellappa, Rama and others},
  booktitle={2024 IEEE International Conference on Robotics and Automation (ICRA)},
  pages={5021--5028},
  year={2024},
  organization={IEEE}
}

@inproceedings{yue2025does,
 author = {Chen, Zhiqi and Lu, Rui and Zhao, Andrew and Wang, Zhaokai and Yue, Yang and Song, Shiji and Huang, Gao},
 booktitle = {Advances in Neural Information Processing Systems},
 editor = {D. Belgrave and C. Zhang and H. Lin and L. Montoya and R. Pascanu and P. Koniusz and M. Ghassemi and N. Chen},
 pages = {57654--57689},
 publisher = {Curran Associates, Inc.},
 title = {Does Reinforcement Learning Really Incentivize Reasoning Capacity in LLMs Beyond the Base Model?},
 url = {https://proceedings.neurips.cc/paper_files/paper/2025/file/537d5aa768c2d534016a4d06f87bc8fb-Paper-Conference.pdf},
 volume = {38},
 year = {2025}
}

@misc{skobelev2023openworlds,
  author = {Vasiliy Skobelev},
  title = {How to create engaging open worlds},
  year = {2023},
  month = {Jul},
  howpublished = {\url{https://web.archive.org/web/20250828213832/https://www.gamesindustry.biz/how-to-create-engaging-open-worlds}},
  note = {Published by GamesIndustry.biz. Archived from the original on Aug 28, 2025}
}

@article{minderer2023scaling,
  title={Scaling open-vocabulary object detection},
  author={Minderer, Matthias and Gritsenko, Alexey and Houlsby, Neil},
  journal={Advances in Neural Information Processing Systems},
  volume={36},
  pages={72983--73007},
  year={2023}
}

@inproceedings{werby2024hierarchical,
  title={Hierarchical open-vocabulary 3d scene graphs for language-grounded robot navigation},
  author={Werby, Abdelrhman and Huang, Chenguang and B{\"u}chner, Martin and Valada, Abhinav and Burgard, Wolfram},
  booktitle={First Workshop on Vision-Language Models for Navigation and Manipulation at ICRA 2024},
  year={2024}
}

@article{yan2025dynamic,
  title={Dynamic open-vocabulary 3d scene graphs for long-term language-guided mobile manipulation},
  author={Yan, Zhijie and Li, Shufei and Wang, Zuoxu and Wu, Lixiu and Wang, Han and Zhu, Jun and Chen, Lijiang and Liu, Jihong},
  journal={IEEE Robotics and Automation Letters},
  year={2025},
  publisher={IEEE}
}

@inproceedings{
    tan2025interactive,
    title={Interactive Post-Training for Vision-Language-Action Models},
    author={Shuhan Tan and Kairan Dou and Yue Zhao and Philipp Kraehenbuehl},
    booktitle={Workshop on Foundation Models Meet Embodied Agents at CVPR 2025},
    year={2025},
    url={https://openreview.net/forum?id=ytGamErI9m}
}

@article{zhou2025beliefmapnav,
  title={Beliefmapnav: 3d voxel-based belief map for zero-shot object navigation},
  author={Zhou, Zibo and Hu, Yue and Zhang, Lingkai and Li, Zonglin and Chen, Siheng},
  journal={arXiv preprint arXiv:2506.06487},
  year={2025}
}

@article{yue2025rl,
  title={RL from Physical Feedback: Aligning Large Motion Models with Humanoid Control},
  author={Yue, Junpeng and Wang, Zepeng and Wang, Yuxuan and Zeng, Weishuai and Wang, Jiangxing and Xu, Xinrun and Zhang, Yu and Zheng, Sipeng and Ding, Ziluo and Lu, Zongqing},
  journal={arXiv preprint arXiv:2506.12769},
  year={2025}
}

@article{gao2025octonav,
  title={OctoNav: Towards Generalist Embodied Navigation},
  author={Gao, Chen and Jin, Liankai and Peng, Xingyu and Zhang, Jiazhao and Deng, Yue and Li, Annan and Wang, He and Liu, Si},
  journal={arXiv preprint arXiv:2506.09839},
  year={2025}
}

@article{wang2026imaginenav++,
  title={ImagineNav++: Prompting Vision-Language Models as Embodied Navigator through Scene Imagination},
  author={Wang, Teng and Zhao, Xinxin and Cai, Wenzhe and Sun, Changyin},
  journal={IEEE Transactions on Pattern Analysis and Machine Intelligence},
  year={2026},
  publisher={IEEE}
}

@inproceedings{wen2025zero,
  title={Zero-shot object navigation with vision-language models reasoning},
  author={Wen, Congcong and Huang, Yisiyuan and Huang, Hao and Huang, Yanjia and Yuan, Shuaihang and Hao, Yu and Lin, Hui and Liu, Yu-Shen and Fang, Yi},
  booktitle={International Conference on Pattern Recognition},
  pages={389--404},
  year={2025},
  organization={Springer}
}

@inproceedings{
    zeng2024poliformer,
    title={PoliFormer: Scaling On-Policy {RL} with Transformers Results in Masterful Navigators},
    author={Kuo-Hao Zeng and Zichen Zhang and Kiana Ehsani and Rose Hendrix and Jordi Salvador and Alvaro Herrasti and Ross Girshick and Aniruddha Kembhavi and Luca Weihs},
    booktitle={8th Annual Conference on Robot Learning},
    year={2024},
    url={https://openreview.net/forum?id=KdVLK0Wo5z}
}

@article{Matterport3D,
  title={Matterport3D: Learning from RGB-D Data in Indoor Environments},
  author={Chang, Angel and Dai, Angela and Funkhouser, Thomas and Halber, Maciej and Niessner, Matthias and Savva, Manolis and Song, Shuran and Zeng, Andy and Zhang, Yinda},
  journal={International Conference on 3D Vision (3DV)},
  year={2017}
}

@inproceedings{savva2019habitat,
  title={Habitat: A platform for embodied ai research},
  author={Savva, Manolis and Kadian, Abhishek and Maksymets, Oleksandr and Zhao, Yili and Wijmans, Erik and Jain, Bhavana and Straub, Julian and Liu, Jia and Koltun, Vladlen and Malik, Jitendra and others},
  booktitle={Proceedings of the IEEE/CVF international conference on computer vision},
  pages={9339--9347},
  year={2019}
}

@article{ramakrishnan2021habitat,
  title={Habitat-matterport 3d dataset (hm3d): 1000 large-scale 3d environments for embodied ai},
  author={Ramakrishnan, Santhosh K and Gokaslan, Aaron and Wijmans, Erik and Maksymets, Oleksandr and Clegg, Alex and Turner, John and Undersander, Eric and Galuba, Wojciech and Westbury, Andrew and Chang, Angel X and others},
  journal={arXiv preprint arXiv:2109.08238},
  year={2021}
}

@inproceedings{yadav2023habitat,
  title={Habitat-matterport 3d semantics dataset},
  author={Yadav, Karmesh and Ramrakhya, Ram and Ramakrishnan, Santhosh Kumar and Gervet, Theo and Turner, John and Gokaslan, Aaron and Maestre, Noah and Chang, Angel Xuan and Batra, Dhruv and Savva, Manolis and others},
  booktitle={Proceedings of the IEEE/CVF Conference on Computer Vision and Pattern Recognition},
  pages={4927--4936},
  year={2023}
}

@article{bai2025qwen2,
  title={Qwen2. 5-vl technical report},
  author={Bai, Shuai and Chen, Keqin and Liu, Xuejing and Wang, Jialin and Ge, Wenbin and Song, Sibo and Dang, Kai and Wang, Peng and Wang, Shijie and Tang, Jun and others},
  journal={arXiv preprint arXiv:2502.13923},
  year={2025}
}

@article{hurst2024gpt,
  title={Gpt-4o system card},
  author={Hurst, Aaron and Lerer, Adam and Goucher, Adam P and Perelman, Adam and Ramesh, Aditya and Clark, Aidan and Ostrow, AJ and Welihinda, Akila and Hayes, Alan and Radford, Alec and others},
  journal={arXiv preprint arXiv:2410.21276},
  year={2024}
}

@inproceedings{anwar2025remembr,
  title={Remembr: Building and reasoning over long-horizon spatio-temporal memory for robot navigation},
  author={Anwar, Abrar and Welsh, John and Biswas, Joydeep and Pouya, Soha and Chang, Yan},
  booktitle={2025 IEEE International Conference on Robotics and Automation (ICRA)},
  pages={2838--2845},
  year={2025},
  organization={IEEE}
}

@article{zhang2024uni,
    title={Uni-NaVid: A Video-based Vision-Language-Action Model for Unifying Embodied Navigation Tasks},
    author={Zhang, Jiazhao and Wang, Kunyu and Wang, Shaoan and Li, Minghan and Liu, Haoran and Wei, Songlin and Wang, Zhongyuan and Zhang, Zhizheng and Wang, He},
    journal={Robotics: Science and Systems},
    year={2025}
}

@article{zhu2025mtu,
  title = {Move to Understand a 3D Scene: Bridging Visual Grounding and Exploration for Efficient and Versatile Embodied Navigation},
  author = {Zhu, Ziyu and Wang, Xilin and Li, Yixuan and Zhang, Zhuofan and Ma, Xiaojian and Chen, Yixin and Jia, Baoxiong and Liang, Wei and Yu, Qian and Deng, Zhidong and Huang, Siyuan and Li, Qing},
  journal = {International Conference on Computer Vision (ICCV)},
  year = {2025}  
}

@article{gong2026stairway,
  title={Stairway to Success: An Online Floor-Aware Zero-Shot Object-Goal Navigation Framework via LLM-Driven Coarse-to-Fine Exploration},
  author={Gong, Zeying and Li, Rong and Hu, Tianshuai and Qiu, Ronghe and Kong, Lingdong and Zhang, Lingfeng and Zhao, Guoyang and Ding, Yiyi and Liang, Junwei},
  journal={IEEE Robotics and Automation Letters},
  year={2026},
  publisher={IEEE}
}

@article{zhu2025strive,
  title={Strive: Structured representation integrating vlm reasoning for efficient object navigation},
  author={Zhu, Haokun and Li, Zongtai and Liu, Zhixuan and Wang, Wenshan and Zhang, Ji and Francis, Jonathan and Oh, Jean},
  journal={arXiv preprint arXiv:2505.06729},
  year={2025}
}

@inproceedings{hossain2024toponav,
  title={Toponav: Topological navigation for efficient exploration in sparse reward environments},
  author={Hossain, Jumman and Faridee, Abu-Zaher and Roy, Nirmalya and Freeman, Jade and Gregory, Timothy and Trout, Theron},
  booktitle={2024 IEEE/RSJ International Conference on Intelligent Robots and Systems (IROS)},
  pages={693--700},
  year={2024},
  organization={IEEE}
}

@article{raychaudhuri2025semantic,
  title={Semantic Mapping in Indoor Embodied AI--A Comprehensive Survey and Future Directions},
  author={Raychaudhuri, Sonia and Chang, Angel X},
  journal={arXiv preprint arXiv:2501.05750},
  year={2025}
}

@article{mobile_sam,
  title={Faster Segment Anything: Towards Lightweight SAM for Mobile Applications},
  author={Zhang, Chaoning and Han, Dongshen and Qiao, Yu and Kim, Jung Uk and Bae, Sung-Ho and Lee, Seungkyu and Hong, Choong Seon},
  journal={arXiv preprint arXiv:2306.14289},
  year={2023}
}

@inproceedings{wang2023yolov7,
  title={{YOLOv7}: Trainable bag-of-freebies sets new state-of-the-art for real-time object detectors},
  author={Wang, Chien-Yao and Bochkovskiy, Alexey and Liao, Hong-Yuan Mark},
  booktitle={Proceedings of the IEEE/CVF Conference on Computer Vision and Pattern Recognition (CVPR)},
  year={2023}
}

@article{shao2024deepseekmath,
  title={Deepseekmath: Pushing the limits of mathematical reasoning in open language models},
  author={Shao, Zhihong and Wang, Peiyi and Zhu, Qihao and Xu, Runxin and Song, Junxiao and Bi, Xiao and Zhang, Haowei and Zhang, Mingchuan and Li, YK and Wu, Yang and others},
  journal={arXiv preprint arXiv:2402.03300},
  year={2024}
}

@inproceedings{ramakrishnan2022poni,
  title={Poni: Potential functions for objectgoal navigation with interaction-free learning},
  author={Ramakrishnan, Santhosh Kumar and Chaplot, Devendra Singh and Al-Halah, Ziad and Malik, Jitendra and Grauman, Kristen},
  booktitle={Proceedings of the IEEE/CVF Conference on Computer Vision and Pattern Recognition},
  pages={18890--18900},
  year={2022}
}

@inproceedings{ramrakhya2023pirlnav,
  title={Pirlnav: Pretraining with imitation and rl finetuning for objectnav},
  author={Ramrakhya, Ram and Batra, Dhruv and Wijmans, Erik and Das, Abhishek},
  booktitle={Proceedings of the IEEE/CVF Conference on Computer Vision and Pattern Recognition},
  pages={17896--17906},
  year={2023}
}

@article{xu2022fast,
  title={Fast-lio2: Fast direct lidar-inertial odometry},
  author={Xu, Wei and Cai, Yixi and He, Dongjiao and Lin, Jiarong and Zhang, Fu},
  journal={IEEE Transactions on Robotics},
  volume={38},
  number={4},
  pages={2053--2073},
  year={2022},
  publisher={IEEE}
}

@article{achiam2023gpt,
  title={Gpt-4 technical report},
  author={Achiam, Josh and Adler, Steven and Agarwal, Sandhini and Ahmad, Lama and Akkaya, Ilge and Aleman, Florencia Leoni and Almeida, Diogo and Altenschmidt, Janko and Altman, Sam and Anadkat, Shyamal and others},
  journal={arXiv},
  year={2023}
}

@article{team2023gemini,
  title={Gemini: a family of highly capable multimodal models},
  author={Team, Gemini and Anil, Rohan and Borgeaud, Sebastian and Alayrac, Jean-Baptiste and Yu, Jiahui and Soricut, Radu and Schalkwyk, Johan and Dai, Andrew M and Hauth, Anja and Millican, Katie and others},
  journal={arXiv},
  year={2023}
}

@article{anthropic2024claude,
  title={The claude 3 model family: Opus, sonnet, haiku},
  author={Anthropic, AI},
  journal={Claude-3 Model Card},
  volume={1},
  pages={1},
  year={2024}
}

\newtcolorbox{vlmprompt}[1][]{
    enhanced,
    breakable,                
    colback=gray!5!white,      
    colframe=gray!75!black,    
    fonttitle=\bfseries,       
    boxrule=0.5pt,             
    arc=4pt,                   
    left=5pt, right=5pt, top=5pt, bottom=5pt, 
    title=#1                   
}

\clearpage
\appendix
\section{Related Work}
\label{app:related_work}

\textbf{Object Navigation Strategies }\quad
Recent work has increasingly leveraged prior knowledge of large models for object navigation. Several methods utilize rich modalities, such as videos, as model inputs to enable the direct output of actions~\cite{zhang2024uni, zhu2025mtu, gao2025octonav, wang2026vlingnav}. 
Although these methods achieve strong results, they need massive training for a particular VLM base model, limiting their ability to benefit from the continuous capability growth and broad world knowledge of foundation model updates. 
In contrast, most frontier-based methods improve real-time performance with zero-shot VLMs by offloading navigation between frontiers to low-level planners. Some existing frontier-based methods rely on smaller VLM models~\cite{chang2023goat, yokoyama2024vlfm, busch2025one, zhang2025apexnav}, which limits their high-level semantic understanding. 
While several studies have attempted to address this issue by adopting large models for frontier selection~\cite{zhou2023esc, yu2023l3mvn, shah2023navigation, zhang2025mem2ego}, these approaches still rely on abstracting the environment into certain intermediate memory representations, thereby limiting the model's capabilities. 

\textbf{Memory Representation and Mapping }\quad
Current methods commonly assist navigation by adopting abstract representations such as topological maps~\cite{wu2024voronav, zhou2024navgpt, hossain2024toponav}, text summarizations~\cite{kuang2024openfmnav, cai2024bridging}, scene graphs~\cite{gu2024conceptgraphs, werby2024hierarchical, yan2025dynamic, yin2025unigoal}, BEV (bird's-eye view) maps~\cite{zhang2025mapnav}, multi-layer maps~\cite{zhu2025strive, zhou2025beliefmapnav, gong2026stairway}, and value maps~\cite{huang2024gamap, long2024instructnav, nie2025wmnav}, but these methods essentially utilize memories as text input for VLM, introducing additional memory processing steps and information bottlenecks. 
Several methods utilize neural rendering techniques to reconstruct 3D scenes, synthesizing novel views for VLM input~\cite{lei2025gaussnav, zheng20263dgsnav, wang2026imaginenav++}. However, the issue of information loss in novel view synthesis remains. 
3D-Mem~\cite{yang20253d} utilizes raw snapshots as direct memory inputs, which better preserves the rich semantic information of the environment. Yet 3D-Mem mainly focuses on memory management, while problems such as dense VLM inference (per 1 meter), and heavy reliance on upstream small models for the object type pre-filtering, remain. Our work further extends the idea of utilizing raw observations to the multi-view, multi-floor large scenarios in a more robust sparse waypoint selection framework.

\textbf{Reasoning and Reinforcement Learning for Embodied Tasks }\quad
To enhance navigational reasoning, several methods employ Monte Carlo Tree Search (MCTS) to generate multiple thought trajectories~\cite{wang2023dreamwalker, wen2025zero}.
An alternative direction is reinforcement learning. With recent paradigms like RLVR, some methods use episodic success as reward~\cite{tan2025interactive, yue2025rl}, but this leads to sparse rewards and poor sample efficiency. 
Other studies~\cite{zeng2024poliformer, qi2025vln} use individual action success as rewards. 
However, this necessitates frequent model interventions for high-frequency, step-by-step decisions, sacrificing time required for complex reasoning, such as Chain-of-Thought. 

While existing work has made noticeable progress in navigation strategies, memory representations, and learning paradigms, they face trade-offs with control frequency, memory abstraction limitations, and reward design dilemmas in reinforcement learning. 
To address these challenges, our method employs a sparse selection framework which enables the VLM to perform in-depth reasoning at critical moments while maintaining effective semantic guidance throughout the task.

\section{Framework Details}

\subsection{Terminology Definitions}

We clarify the terminology used throughout the paper and discuss the fairness of our comparisons.

\paragraph{Zero-shot High-level Navigation}
We use ``zero-shot'' to describe the high-level, training-free VLM module of \textit{PIGEON-ZeroShot}, which accounts for the vast majority of the system's parameters. In this setting, the VLM is used in its frozen form and is not fine-tuned on ObjectNav trajectories, simulator rollouts, or benchmark-specific annotations. The VLM receives only the online observations and the constructed PoI candidates, and selects among them through prompting at test time. Nevertheless, due to the lack of an off-the-shelf stair navigation policy, the specific stair navigation module in PIGEON is trained on the HM3Dv1 train split and employed during evaluation across different datasets. Such terminology for zero-shot object navigation is consistent with recent literature, such as VLFM~\citep{yokoyama2024vlfm}, which employs a trained PointNav RNN for low-level execution in the Habitat simulator. This modular architecture is particularly advantageous for real-world applications, as the learned low-level policy can be seamlessly substituted with commercial out-of-the-box PointNav systems or heuristic algorithms such as A* and the Fast Marching Method (FMM).

\paragraph{RLVR-trained PIGEON}
\textit{PIGEON-RLVR} is not training-free. It fine-tunes the high-level VLM selector using automatically constructed verifiable rewards derived from simulator trajectories. We therefore report \textit{PIGEON-ZeroShot} and \textit{PIGEON-RLVR} separately.  \textit{PIGEON-ZeroShot} evaluates the effectiveness of the PoI interface with a frozen VLM, while \textit{PIGEON-RLVR} evaluates whether the same interface enables scalable reinforcement learning from verifiable navigation feedback.

\paragraph{Oracle-Free Evaluation}
At test time, PIGEON does not use privileged information such as the ground-truth goal location, oracle shortest path to the target, ground-truth semantic maps, or manually annotated object locations. The agent only observes RGB-D inputs, camera poses, and the outputs of its perception modules. Shortest-path distances or target visibility signals are used only for constructing rewards during RLVR training or for evaluation, not for test-time decision making.

\paragraph{Summary}
In summary, our claims are scoped as follows.  \textit{PIGEON-ZeroShot} is zero-shot and training-free with respect to the high-level decision module, but it uses pretrained execution modules, which aligns with common practices in recent zero-shot ObjectNav frameworks. \textit{PIGEON-RLVR} is a trained variant that fine-tunes the high-level selector using automatically verifiable rewards. Neither variant uses test-time oracle goal information or ground-truth semantic maps.

\subsection{PoI Concept}
A traditional Point of Interest in geographic information systems is a data unit used to mark specific locations, representing entities such as buildings, stores, and bus stops. It is one of the basic types of geospatial data that people might be interested in for particular reasons. Game designers propose similar concepts in open-world game industry, where PoIs are defined as sparse raw semantic landmarks in large maps to guide the exploration of the players \cite{skobelev2023openworlds}. Similarly, we formulate the whole navigation process as analyzing and navigating between PoIs in order to guide the exploration of the robots for regions they might be interested in. Our approach models navigation as a sparse selection problem, such that the input of VLM is mainly restricted to raw observations and its output is limited to the given PoI options. This design eliminates excessive reliance on intermediate representations during processing, thereby reducing its hallucination issues during inference. 

\subsection{PoI Design Details}

In this section, we detail the generation and utilization of various PoIs, which serve as critical waypoints for the VLM-guided agent.

\textbf{Frontier PoI}\quad If an observation $O_t$ expands the explored area, a non-directional PoI on the boundary of the newly expanded area is created. If the frontier PoI is no longer on the exploration frontier, or it has been selected, it will be removed from the PoI set. As the scale of areas and number of floors in a map increase, a few raw observations may not be sufficient to capture the full semantic information of the environment. We used the MCP algorithm to augment Frontier PoI selections. The memory complement strategy aims to enrich the environment semantics surrounding targeted PoIs, which unleash the power of increasingly powerful VLM models furthermore. PoI markers are then projected onto 2D observations, and assigned with a unique number. For each PoI $p_i$, we use the camera parameters to project the PoI's map location into the coordinate system of its corresponding image. A circular numeric marker with the number $i$ is then overlaid on the image. If the marker of the PoI is on the edge of the observation, then the PoI will be filtered out. Otherwise, $n_{prox}$ nearest PoIs are collected and fed into the VLM for selection based on the consideration of proximity.

\textbf{Suspect Object PoI}\quad To address the issue of confidence distortion in the detector, we use a VLM to validate the current observation when the confidence lies between $\tau_{sus}$ and $\tau_{target}$. Based on the VLM’s feedback, we multiply the confidence of the corresponding 3D object by a coefficient. Furthermore, We generate Suspect Object PoIs around the object, enabling the robot to actively move to the object’s vicinity and observe it from different angles with multi-view image observations. Suspect object viewpoint PoI has a higher priority than normal frontiers. Whenever the agent reaches a PoI around the potential target, the agent will turn to face the object, and its image observation is added into $\mathcal{S}_x$. The images within $\mathcal{S}_x$ are then concatenated and passed to the VLM to confirm the identity of the suspected object. For computational efficiency, the total number of object confirmations by the VLM does not exceed $t_{con}$, the total number of active confirmations does not exceed $t_{acon}$, and the VLM performs unified confirmation for all objects in the frame.

\textbf{Stair PoI}\quad During exploration, a height map is updated by transforming the depth image $d_t$ into 3D point cloud, and projecting the point cloud onto a BEV grid map. Points that exceed the robot's height are filtered out, and the highest of the remaining points becomes the height of that grid. Grids whose height is lower than $h_{floor}$ are recognized as a sinking region. To enable multi-layer exploration, the staircases are regarded as a special object category. After the confidence of a staircase exceeds $\tau_{stair}$, VLM examines the current RGB image and report the stair type if visible. If it leads upstairs or downstairs in a sinking region, the edge of stair region with several observation images will be integrated into a Stair PoI. For floor navigation, we identify staircases and model their passable regions as frontiers during navigation process, thereby extending the frontier selection paradigm to include floor transitions. We trained a floor navigation model similar to the PointNav model used for ground-level navigation.

\textbf{Memory Complement}\quad MCP have been proved as a NP-hard problem \cite{feige1998threshold}, thus we use the greedy algorithm for the MCP which has provided a well-known approximation guarantee of $1 - 1/e$ \cite{nemhauser1978analysis} in comparison of the optimal solution. BEV frustums used during the algorithm are re-computed based on the current obstacle distribution.

\textbf{Floor Transition}\quad The floor transition procedure consists of three parts. Firstly, the frontier boundary between exploration area and every upstairs and downstairs region will be calculated, and ranked based on the area size which the boundary belongs to. After that, Stair PoIs is generated from top $n_{stair}$ ranked boundary frontiers, with the observations added by the memory complement strategy. Secondly, VLM select Stair PoIs based on the observation, and the selected PoI is marked as the navigation waypoint. Lastly, after the robot reaches the stair PoI, The robot will go up and down the stairs. When going down the stairs, the robot will execute LOOK\_DOWN command and constantly moving forward into the sinking area using PointNav RNN until reaching a flat plane. When going up the stairs, we trained the StairNav RNN for action execution.

\textbf{Stagnation Detection} \quad Upstream obstacle map and detection modules causes the frontier to appear in areas that are actually impassable, or results in objects unrelated to the navigation target with a high confidence level. As the map expands, an excessive number of PoIs may appear on a single floor, preventing the robot from quickly navigating to distant but semantically relevant areas. Therefore, in addition to using VLM to select PoIs, we have introduced a PoI deactivation mechanism based on the memory complement strategy. When a robot lingers in a particular area for too long, PIGEON retrieves relevant historical images using the MCP algorithm and passes them to the VLM to determine whether the robot should leave that area. Once the steps navigating to the target PoI exceeds $t_{stuck}$, we retrieve the reference images whose frustum covers the PoI’s location. If there is more than a single reference image, we feed the reference images after MCP filtration into the VLM. We also input a color BEV map containing the robot’s trajectory, where locations VLM lingering are marked larger. If the VLM determines that further navigation to the PoI is unnecessary, or insufficient image coverage indicates the robot is stuck, the area within a certain radius $r_{ban}$ around the PoI will be marked as impassable. Subsequently, any PoIs generated within this area will not be selected by the VLM.

\subsection{A-EQA Design}

Our A-EQA implementation extends the ObjectNav pipeline while preserving its navigation backbone. In both tasks, the agent constructs an obstacle-aware BEV map from RGB-D observations, extracts navigable PoIs, uses a VLM to select promising exploration directions, and relies on a PointNav policy for local motion. The two variants also share temporary banning of stagnant goals and VLM-guided floor transitions through staircases.

The key difference is the task objective. In ObjectNav, the input is an object category, such as chair. Once a reliable target instance is detected and confirmed, the planner switches from exploration to target pursuit, navigates toward the closest point on the object point cloud, and stops when it reaches the object vicinity.

For EQA, the input is a natural-language question, such as “Is the ceiling fan on?”. A VLM first extracts a detector-friendly core object, such as ceiling fan, which is used to guide the navigation process. However, the complete question remains the semantic objective for subsequent VLM queries. Frontier PoIs, Object PoIs, Stagnant PoI reviews, Floor Summary PoIs, and Stair PoIs management are reformulated as evidence-seeking decisions. The agent chooses waypoints that are likely to reveal sufficient visual evidence for answering the question.

Unlike ObjectNav, detecting the core object is not a terminal condition in A-EQA. A confirmed detection may identify a useful region or viewpoint, but the planner does not enter the ObjectNav Goal-Navigation phase. Instead, it continues active exploration until a VLM call returns a non-empty answer enclosed by an \texttt{[EQA]...[/EQA]} tag. The agent then stops immediately and records the generated answer. If no answer is obtained within the episode step limits, it returns a fallback response stating that sufficient visual evidence could not be found.

A-EQA episodes also differ at the dataset and evaluation levels. The A-EQA dataset provides a question, an initial pose, and a reference answer, but no physical navigation goal. We therefore wrap each A-EQA record as a Habitat ObjectNav-compatible episode using synthetic goal viewpoints solely for simulator compatibility. Since these synthetic goals do not represent the task objective, ObjectNav-specific path-efficiency measures such as SPL and SoftSPL are disabled. Instead, the implementation saves generated answers, extracted core targets, reference answers, and traveled path lengths for A-EQA evaluation, which is performed by an LLM evaluator as in \citep{yang20253d}.

\section{Details on RL}
Our design enables the VLM to be fine-tuned directly using RLVR (Reinforcement Learning with Verifiable Rewards). Due to the difference of training distribution between vision-language tasks and indoor navigation, we employ RLVR to enhance PoI selection, allowing the targeted VLM model to align with the navigation distribution, developing sophisticated reasoning for complex scenarios without expensive annotation.
\label{app:rl}
\subsection{RLVR Design}
Our method can be fine-tuned using reinforcement learning on any navigation dataset that provides BEV 2D maps and target object annotations, enhancing its reasoning capabilities without requiring explicit Chain-of-Thought data. 

The reinforcement learning dataset consists of tuples $(v, \mathcal{T}, y)$, where $v$ is the image inputs for the VLM, $\mathcal{T}$ is the instruction, while data caption $y$ varies according to the question type. For Frontier PoIs selection, $y=\mathbf{D}$, a list of distances from each candidate PoI to the target objects. These distances are computed using a ground truth path planner on the BEV map. For object confirmation, $\mathrm{y}=N$, indicating the number of points the target object shown in $v$. The model is required to output \texttt{<think> A </think> <answer> B </answer>}, where A represents chain-of-thought and B represents the target PoI ID or confirmation decision.  

During an episode, after arriving at a PoI waypoint $\mathbf{p}_k$, the agent follows an distance-aware $\epsilon$-greedy policy for data collection, in order to enhance the model's capability with suboptimal decision history. Rather than using a fixed exploration rate, the probability of selecting the optimal PoI (i.e., the greedy action with probability $1-\epsilon$) is dynamically adjusted based on the navigation distance to the nearest goal, denoted as $d_{goal}$. If a sub-optimal action is triggered (with probability $\epsilon$), the probability mass is uniformly distributed among the remaining $n-1$ candidates, where $n$ is the total number of candidate PoIs. This design ensures higher randomness when the robot is far from the goal, while encouraging optimal exploitation as it approaches the target. 
With probability $1-\epsilon$, it selects the PoI $P_j$ corresponding to the minimum distance $d_j = \min(\mathbf{D})$. Otherwise, it selects a random PoI. The 2D location $p_j$ of the chosen PoI is then set as the next waypoint $\mathbf{p}_{k+1}$.

To avoid the need for manual CoT annotations, we employ the Group Relative Policy Optimization (GRPO) from \cite{shao2024deepseekmath} as our RLVR algorithm. For a given input $q=(v, \mathcal{T})$ and an output token $o_{i,t}$ in group roll-out number $i$, we extract chosen PoI index $j$ from response $o_i$ and use the distance list $\mathbf{D}=(d_1,\dots, d_j,\dots, d_n)$ to compute its reward $R_i$. Consider the current VLM policy $\pi_\theta$ and the reference policy from the previous iteration $\pi_{old}$, the importance weight is $W_{i,t} = \pi_\theta(o_{i,t}|q, o_{i,<t}) / \pi_{old}(o_{i,t}|q, o_{i,<t})$. We optimize the policy by maximizing the following objective function over a group of $G$ sampled responses:
\begin{equation}
\begin{aligned}
\mathcal{J}(\theta)&=\mathrm{E}[\{o_i\}_{i=1}^G\sim \pi_{old}]\frac{1}{G}\sum_{i=1}^G \frac{1}{|o_i|}\sum_{t=1}^{|o_i|}\{\min(W_{i,t}A_{i}, \mathrm{clip}(W_{i,t})A_{i})-\beta\mathrm{KL}[\pi_\theta \Vert \pi_{ref}]\}
\end{aligned}
\end{equation}
where $o_i$ is the $i$-th sampled response from the VLM, and $A_i$ is its relative advantage over other samples in the group, which is computed via normalization as in \eqref{reward_design}.

\subsection{More RLVR Analysis}

\begin{table}[htbp]
\caption{RLVR Model Evaluation Metrics on the whole HM3Dv2 dataset.}
\small
\label{tab:metrics_table}
\centering
\begin{tabular}{lcccccc}
\toprule
Configuration & SR$\uparrow$ & SPL$\uparrow$ & Entropy & Avg Len & \# Out=0 & \# Unconf. \\ 
\midrule
vanilla 7b & 72.66 & 33.16 & 0.5205 & 53.67 & 92 & 94 \\
\midrule
\textbf{w} format reward & 73.00 & 33.71 & 0.8466 & 94.85 & 182 & 93 \\
\textbf{w} selection reward & 73.42 & 34.67 & 0.9402 & 101.16 & 31 & 172 \\
\textbf{w} confirmation reward & 73.80 & 34.15 & 0.8971 & 97.49 & 209 & 48 \\
\midrule
\textbf{w} all reward & 73.50 & 35.02 & 0.9075 & 78.02 & 5 & 2 \\
\textbf{w} online filtering & 74.00 & 34.99 & 0.9615 & 76.14 & 10 & 3 \\
\textbf{w} gpsize = 10 & 74.40 & 35.14 & 0.9396 & 74.59 & 12 & 1 \\
\textbf{w} step = 1000 & 75.00 & 35.52 & 0.9290 & 91.00 & 0 & 57 \\
\textbf{w} large scale & 75.72 & 36.99 & 0.7779 & 81.28 & 3 & 58 \\
\midrule 
\textbf{w} 30k data & 76.90 & 36.56 & 0.9983 & 81.58 & 0 & 34 \\
\textbf{w/o} KL & 77.56 & 36.55 & 1.4393 & 86.85 & 0 & 0 \\
\bottomrule
\end{tabular}
\end{table}

We present our complete RLVR results in Table \ref{tab:metrics_table}. Notably, while overfitting on small datasets leads to a decrease in entropy, leveraging high-quality data and removing the KL penalty significantly increases it~\citep{li2026questa}. We hypothesize that this increased entropy reflects the policy distribution converging closer to an ideal, robust navigation strategy. Furthermore, relying exclusively on the Format Reward induces a pathological increase in output length. We attribute this to reward exploitation. The model bypasses output correctness and simply minimizes the probability of format errors by generating excessively long responses. Comprehensive hyper-parameters regarding the RLVR training environment are provided in Appendix \ref{rlvr_training_details}.

\begin{table}[h]
\caption{Ablation study on the RL module using Qwen-2.5VL-7B on a subset of HM3Dv2.}
\small
\centering
\begin{tabular}{@{}lcccc@{}}
\toprule
\textbf{Method} & \textbf{SR}$\uparrow$ & \textbf{SPL}$\uparrow$ & \textbf{Soft-SPL}$\uparrow$ & \textbf{Avg. Time}$\downarrow$ \\
\midrule
Qwen-2.5VL-7B(Base) & 74.7 & 30.2 & 31.4 & 85.62s \\
+ Confirm Reward & 75.8 & 28.7 & 29.9 & 85.70s \\
+ Binary Reward & 75.6 & 31.6 & 33.0 & 82.67s \\
+ Soft Reward & 75.8 & 33.2 & 34.6 & 80.02s \\
\bottomrule
\end{tabular}
\label{tab:ablation_rl}
\end{table}

We also conduct an experiment on the reward types for PoI selections in Table~\ref{tab:ablation_rl}. Simply using a random reward for the VLM's PoI selections (+ Confirm Reward) results in a significant drop in SPL, which demonstrates that PoI selection primarily enhances navigation efficiency. 
Using a binary 0-1 reward based on whether the selected PoI is closest to the target improves the model's capabilities, but less effectively than using a soft reward. 
Compared to the base Qwen model, applying our soft reward as in \eqref{reward_design} increases the SR by 1.1\% and the SPL by 3.0\%, demonstrating an improvement in navigation efficiency.

\subsection{Details on StairNav Training}
\textbf{StairNav Dataset Construction}\quad To train the StairNav RNN policy, we construct a PointNav dataset for inter-floor navigation based on HM3D train split. Specifically, we extract stair regions from HM3D semantic annotations. Since HM3D stair annotations are noisy (e.g., a single staircase may be split into multiple segments, multiple staircases may be merged, and some annotated stairs are too short), directly using them may cause premature stopping or failure to stop after traversing multiple floors. To address this, we first sample navigable points in each scene and apply DBSCAN to estimate floor elevations. We then sample navigable points near stairs on adjacent floors and use each such pair as the start and goal locations for StairNav, with a randomly sampled initial orientation.

\textbf{Model Structure and Training}\quad Since StairNav and PointNav share many task characteristics, we initialize the StairNav policy with the weights of the PointNav policy from VLFM to leverage the priors learned from PointNav. The StairNav policy uses ResNet-18 as the backbone and a two-layer LSTM recurrent module. To prevent training collapse, we freeze the vision encoder. We train the policy using Variable Experience Rollout (VER)~\citep{wijmans2022ver} with the following hyperparameters: value loss coefficient $=0.5$, entropy coefficient $=0.01$, learning rate$=2.5\times10^{-4}$, $\varepsilon=10^{-5}$, discount factor $\gamma=0.99$, and $\tau=0.95$.

\section{More Experiment Results}
\label{app:real_world}

\subsection{Details on Real-World Deployment}

\textbf{Unitree Go2 Quadrupedal Robot}\quad The physical test environment mainly consists of three areas: an office area, a pantry, and a three-bedroom apartment. The apartment further includes a living room, bedrooms, and bathrooms. The robot is equipped with a 3D Lidar used for low-level localization and navigation, and an RealSense D455 camera for object detection. Due to the inaccuracy of depth camera, we run a lidar-SLAM system \cite{xu2022fast} to provide the pose of robot and a 2D occupancy map in real-time. The model and ROS main node is hosted on a single laptop with a RTX 5090 Laptop GPU.

\begin{figure}[h]
\centering
\includegraphics[width=0.7\linewidth]{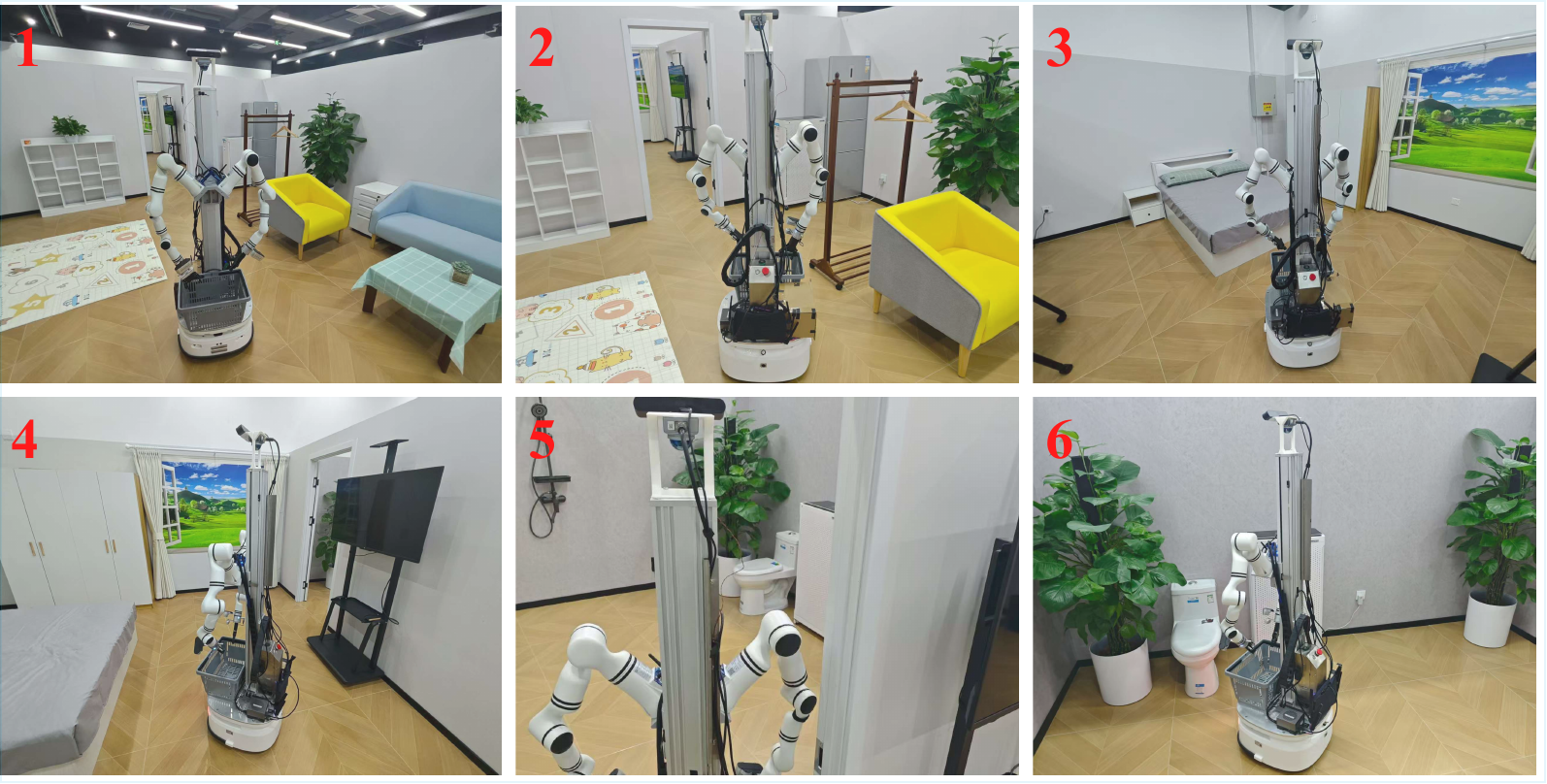} 
\caption{Wheeled robot experiment in the physical environment finding for a toilet.}
\label{realworld_old}
\end{figure}

\textbf{Wheeled Robot}\quad We also validate the proposed method on a wheeled robot as shown in Figure \ref{realworld_old}. The robot is equipped with a 3D Lidar used for low-level localization and navigation, and an RGBD camera for object detection.

To evaluate the practicability of our model, we conduct real-world robot experiments. Shaking during quadruped robot locomotion causes motion blur in the captured images, which increases the variance of the experimental results. Therefore, we use a wheeled robot with more stable motion to evaluate the model’s real-world performance as shown in Table \ref{table4}, with 20 episodes in total and various objects as our goal objects. We adopt quantized \textit{PIGEON-7B} as an \textit{PIGEON-RLVR} vision based on Qwen-2.5VL-7B after RLVR for our method, quantized Qwen-2.5VL-7B for InstructNav, and preserve the original BLIP2 for VLFM which utilizes a smaller VLM model. The intermediate processing pipeline of our model, including PoIs generation, semantic map construction, detection, etc., achieves an inference frequency of 2Hz.  \textit{PIGEON-7B} demonstrates better success rate and lower execution time among other methods, showing both the generalization ability and time advantage in real-world scenarios.

\begin{table}[t]
\small
\centering
\caption{Comparison with other method in terms of testing time, number of VLM calls and SR in the real-world setting.}
\setlength{\tabcolsep}{2mm}
\begin{tabular}{@{}lcccc@{}}
\toprule
\textbf{Method} & \textbf{VLM Type} & \textbf{SR}$\uparrow$ & \textbf{Avg. Time}$\downarrow$ & \textbf{Avg. VLM Calls}$\downarrow$ \\
\midrule
\textit{PIGEON-7B} & \text{Qwen} & 95\% & 48.91s & 6.2 \\
VLFM & \text{BLIP} & 85\% & 49.25s & 53.3 \\
InstructNav & \text{Qwen} & 60\% & 87.83s & 23.8 \\
\bottomrule
\end{tabular}
\label{table4}
\end{table}

\subsection{Further Analysis for Simulation}

\textbf{Further Analysis for Main Results}\quad For our non-RL approach \textit{PIGEON-ZeroShot}, we use GPT-4o-2024-02-15-preview \cite{hurst2024gpt} as the VLM and compare against several state-of-the-art methods. In Table \ref{tab:performance}, various methods are classified based on whether they require training, whether they require the simulator to provide oracle information beyond RGB-D images and odometry, and the form of the intermediate information representation passed to the main large model. We classify methods free from training and oracles as zero-shot methods, due to their independence from specific data sources.

Intermediate representation is defined as the input of the main neural network for models requiring training, and the input of the LLM or VLM for methods without training. PONI uses semantic BEV maps as the input of neural networks, while PIRLNav and Uni-Navid utilize implicit history token for the Transformer architecture. Methods like STRIVE or UniGoal adopt multi-region topological maps or 3D scene graphs in the navigation process. Essentially, these abstract representations serve as structured texts sent into large models. VLFM and ApexNav leverage raw images as the input of BLIP-2 for text-image matching, while WMNav and ImagineNav++* employ raw images as the input of VLM. However, ImagineNav++ needs observations from novel view angles for practical deployment, thus Novel View Synthesis(NVS) is introduced for VLM input.

We provide a simulation demonstration as follow.

\begin{figure}[h]
\centering
\includegraphics[width=\linewidth]{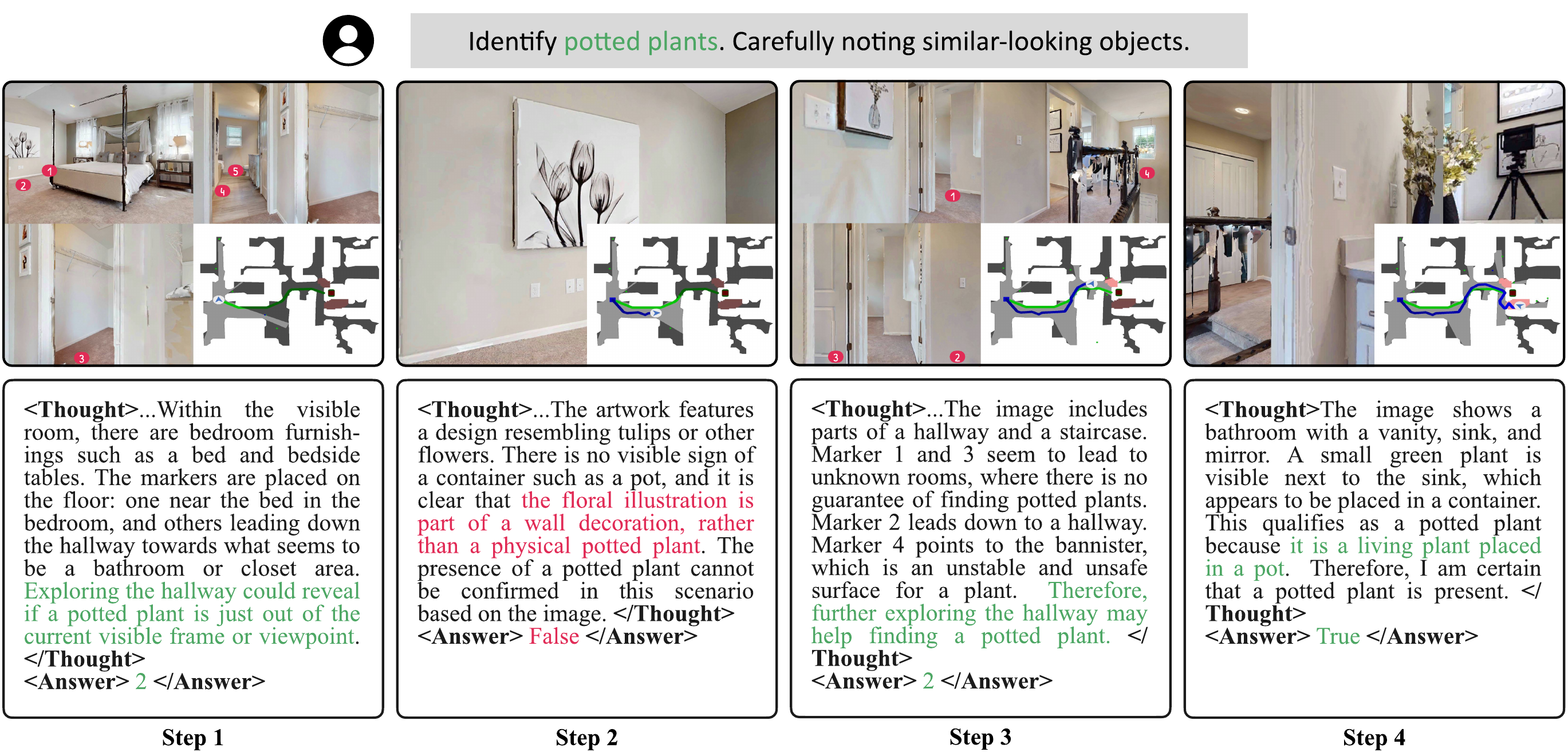} 
\caption{Simulation demonstration in HM3D dataset. At step 1, PIGEON spins around and determines mark 2 to be the most promising PoI. At step 2, PIGEON encounters an artwork of flower. The low level detector identifies the artwork as a potted plant, yet VLM denies it by exploiting information from environment. At step 3, PIGEON determines mark 2 to be the most promising PoI. At step 4, PIGEON confirms that it is facing a real potted plant, and calls stop.}
\label{simulation}
\end{figure}

\textbf{Further Analysis for Parallel Evaluation}\quad To speed up the evaluation process, we employed a parallel evaluation approach when testing on the simulator. Table \ref{tab:parallel_performance} shows the relationship between evaluation time and the number of evaluation threads, demonstrating our method can be deployed multiple times on a single RTX 4090. Furthermore, based on our local deployment experiments, deploying our method encounters no performance issues even on an older graphics card, e.g. GTX 1070.

\begin{table}[htbp]
    \centering
    \caption{Performance and time of our method during parallel evaluation on a subset of HM3Dv2.}
    \label{tab:parallel_performance}
    \begin{tabular}{lcccccc}
        \toprule
        Parallel Cnt & 1 & 2 & 4 & 8 & 12 & 16 \\ 
        \midrule
        SR & 75.00 & 75.00 & 79.00 & 74.00 & 73.00 & 78.00 \\ 
        SPL & 37.92 & 38.87 & 40.17 & 38.55 & 38.49 & 37.27 \\ 
        Avg time (sec) & 163.41 & 166.02 & 179.48 & 214.98 & 268.38 & 349.45 \\ 
        \bottomrule
    \end{tabular}
\end{table}

\section{Experiment settings}

\subsection{VLM Prompts for Object Navigation}

\begin{vlmprompt}[Frontier PoI Selection]
\textbf{User:} You are a robot and after exploration, you have the given images.
You are given first-person observations of candidate frontiers.
The same numbered marker can appear in multiple images and always refers to the same candidate frontier.
Some candidates may have fewer observations than others.
Each candidate view represents a possible direction to find the \{goal\_object\}.
Based on the numerical markers in the images, select one of these numbers to move next.
If you're not confident in moving to the marker to find the \{goal\_object\}, you should return \texttt{<answer> 0 </answer>}.
Please note all closed doors cannot be opened.
Please don't select frontiers leading to dark regions or dark hallways which lack data.
Please select frontiers which is on the ground rather than those floating in air.
Please follow the format like this, without bracket -- \\
\texttt{<thought> [put your step-by-step thinking process here] </thought>} \\
\texttt{<answer> [put a single marker id or 0 here] </answer>}
\end{vlmprompt}

\begin{vlmprompt}[Candidate Object Generation]
\textbf{System:} You are an intelligent agent that helps the object detector identify detected objects. 
Based on the labels I provide, you'll give the labels that may cause misdetection, potential room objects, and the confidence threshold.
At each step, I will give you a label. You should consider the following:
(1) For misdetection, provide objects that may be misdetected by detectors such as YOLOv7, considering similarities in shape, color, size, etc. The labels should be in COCO classes, with at least 3 and at most 5 labels.
(2) For the confidence score, the confidence score should be in [0.25, 0.65], as the smallest is an apple in a home (0.25) and the largest is a bed in a home (0.65).
Please follow the format like this, without bracket -- \\
\texttt{<thought> [put your step-by-step thinking process here] </thought>} \\
\texttt{<answer> [put a list of objects and a confidence score here] </answer>}

\textbf{User:} cake

\textbf{Assistant:} 
\texttt{<thought>} In the COCO classes, confidence threshold for cake: A cake is typically round or square and can have decorative icing or layers. It is quite smaller than a bed (0.65), but bigger than an apple (0.25), so the confidence threshold should be low, around 0.30, which is in [0.25, 0.65].
Misdetection possibilities:
donut: A donut, especially a large one or a stack of donuts, can resemble a cake because of its circular shape and icing-like decorations.
pizza: A whole pizza, particularly one with thick crust or layered toppings, might resemble a cake in certain contexts.
sandwich: A stacked sandwich, especially one with visible layers, could look similar to a layered cake.
pie: A pie can be misdetected as a cake due to its circular shape and pastry-like crust, which is similar to a cake's appearance.\texttt{</thought>} \\
\texttt{<answer> [donut, pizza, sandwich, pie, 0.30] </answer>}

\textbf{User:} cabinet

\textbf{Assistant:} 
\texttt{<thought>} In the COCO classes, confidence threshold for cabinet: A cabinet is typically rectangular, with flat surfaces, often in the form of a box or with shelves inside. It's smaller than larger furniture pieces like beds or dining tables (0.60) but still sizable compared to smaller household objects. The confidence threshold for a cabinet should be around 0.50, which is in [0.25, 0.65].
Misdetection possibilities:
bookshelf: A bookshelf can be similar in shape and size to a cabinet, with a flat surface and storage space, which may lead to confusion.
dresser: A dresser, especially one with drawers, could resemble a cabinet because of the similar rectangular shape and flat surfaces.
closet: A closet, particularly a small one, can resemble a cabinet in terms of its rectangular form and purpose for storage. \texttt{</thought>} \\
\texttt{<answer> [bookshelf, dresser, closet, 0.50] </answer>}

\textbf{User:} \{goal\_object\}
\end{vlmprompt}

\begin{vlmprompt}[Object PoI Confirmation]
\textbf{User:} You are an image recognition assistant. You will be presented with an object's image captured from single or multiple views. Your task is to check whether the \{goal\_object\} is present in the given image. Please do not confuse \{goal\_object\} with the following objects: \{object\_list\}. Only respond "True" if you are certain the \{goal\_object\} is clearly present in the image and located indoors. Respond "False" if the target object is clearly not present. Respond "Unconfirmed" if you are unsure whether the target object is present, or if its location cannot be confidently determined. Respond "Outdoor" if it is present but located outdoors.
Please follow the format like this, without bracket -- \\
\texttt{<thought> [put your step-by-step thinking process here] </thought>} \\
\texttt{<answer> [put True, False, Unconfirmed or Outdoor here] </answer>}
\end{vlmprompt}

\begin{vlmprompt}[Stair PoI Selection]
\textbf{User:} You are helping an indoor robot choose a navigation frontier near a stair.
You are given first-person observations of up to 4 candidate frontiers.
The same numbered marker can appear in multiple images and always refers to the same candidate frontier.
Some candidates may have fewer observations than others.
The semantic search target is \{goal\_object\}.
Select the best numbered frontier that is near the stair.
If none looks reliable, return \texttt{<answer> 0 </answer>}.
Please follow the format: \\
\texttt{<thought> [reasoning content] </thought>} \\
\texttt{<answer> [marker id or 0] </answer>}
\end{vlmprompt}

\begin{vlmprompt}[Stagnant Detection]
\textbf{User:} You are helping an indoor robot decide whether to temporarily ban its current navigation goal.
Image 1 is the current BEV map.
Image 2 is the current color BEV map.
Images 3 to 6 are recent first-person views related to the current goal.
The robot has been stagnant near this goal.
The semantic search target is \{goal\_object\}.
When visible, the same current goal is marked in the first-person views by a red circular marker labeled 1.
Reply 1 if this goal region should be banned, otherwise reply 0.
The goal is a red point on the BEV and colored BEV map.
If the red-point goal seems irrelevant to the semantic target or clearly unreachable, ban the goal.
If the red-point goal seems irrelevant to the target question or clearly unreachable, ban the goal.
Please follow the format: \\
\texttt{<thought> [reasoning content] </thought>} \\
\texttt{<answer> [1 or 0] </answer>} \\
\end{vlmprompt}

\begin{vlmprompt}[Floor Summary PoI Confirmation]
\textbf{User:} You are helping an indoor robot decide whether it should leave the current floor and search via stairs.
Images 1 to 6 are first-person views selected to summarize covered observations on the current floor.
The semantic search target is \{goal\_object\}.
Reply 1 if the robot should go search on another floor now, otherwise reply 0.
Observations leading to dark regions or dark hallways which lacks data don't represent the environment undiscovered.
Even a short flight of stairs can create a barrier between different levels
Please follow the format: \\
\texttt{<thought> [reasoning content] </thought>} \\
\texttt{<answer> [1 or 0] </answer>}
\end{vlmprompt}

\subsection{VLM Prompts for Active-EQA}

\begin{vlmprompt}[Question Object Extraction]
\textbf{User:} You help an indoor robot answer embodied QA questions.
Given the episode question, extract the single core physical object or place that the robot should search for first.
Use a short detector-friendly noun phrase, for example: ceiling fan, bed, painting, table, sink, refrigerator, door, clock.
If the question is about a room-level property, return the most relevant room or object.
Please follow the format: \\
\texttt{<thought> [brief reason] </thought>} \\
\texttt{<answer> [core object noun phrase] </answer>} \\
Episode question: {question}
\end{vlmprompt}

\begin{vlmprompt}[Frontier PoI Selection]
\textbf{User:} You are a robot and after exploration, you have the given images.
You are given first-person observations of candidate frontiers.
The same numbered marker can appear in multiple images and always refers to the same candidate frontier.
Some candidates may have fewer observations than others.
The target question is: \{goal\_object\}
If any candidate image contains enough visual evidence to answer the target question, include the concise answer as \texttt{[EQA]answer[/EQA]}.
Each candidate view represents a possible direction to find evidence for answering the target question.
Based on the numerical markers in the images, select one of these numbers to move next.
If you're not confident in moving to the marker to answer the target question, you should return \texttt{<answer> 0 </answer>}.
Please note all closed doors cannot be opened.
Please don't select frontiers leading to dark regions or dark hallways which lacks data.
Please select frontiers which are on the ground rather than those floating in air.
Please follow the format like this, without bracket -- \\
\texttt{<thought> [put your step-by-step thinking process here] </thought>} \\
\texttt{<answer> [put a single marker id or 0 here] </answer>} \\
\texttt{[EQA][only if answerable, concise QA answer][/EQA]}
\end{vlmprompt}

\begin{vlmprompt}[Object PoI Confirmation]
\textbf{User:} You are an image recognition assistant helping an indoor robot answer an embodied QA question.
The target question is: \{goal\_object\}
You will be presented with an object's image captured from single or multiple views.
First, decide whether the image(s) contain enough visual evidence to answer the target question.
If they do, include a concise answer in the exact form \texttt{[EQA]answer[/EQA]}.
Always include an \texttt{[EQA]...[/EQA]} line; leave it empty as \texttt{[EQA][/EQA]} if the image(s) do not answer the question.
Also check whether the visual evidence is relevant to the target question.
Please do not confuse the needed evidence with these related detector classes: \{object\_list\}.
Respond \texttt{<answer> True </answer>} if the needed evidence is clearly present indoors,
\texttt{<answer> False </answer>} if it is clearly not present,
\texttt{<answer> Unconfirmed </answer>} if unsure, or
\texttt{<answer> Outdoor </answer>} if it is outdoors.
Please follow the format: \\
\texttt{<thought> [brief visual reasoning] </thought>} \\
\texttt{<answer> [True, False, Unconfirmed or Outdoor] </answer>} \\
\texttt{[EQA][concise QA answer, or empty if not answerable][/EQA]}
\end{vlmprompt}

\begin{vlmprompt}[Stair PoI Selection]
\textbf{User:} You are helping an indoor robot choose a navigation frontier near a stair.
You are given first-person observations of up to 4 candidate frontiers.
The same numbered marker can appear in multiple images and always refers to the same candidate frontier.
Some candidates may have fewer observations than others.
The target question is: \{goal\_object\}
If any image contains enough visual evidence to answer the target question, include the concise answer as \texttt{[EQA]answer[/EQA]}.
Select the best numbered frontier that is near the stair.
If none looks reliable, return \texttt{<answer> 0 </answer>}.
Please follow the format: \\
\texttt{<thought> [reasoning content] </thought>} \\
\texttt{<answer> [marker id or 0] </answer>} \\
\texttt{[EQA][only if answerable, concise QA answer][/EQA]}
\end{vlmprompt}

\begin{vlmprompt}[Stagnant Detection]
\textbf{User:} You are helping an indoor robot decide whether to temporarily ban its current navigation goal.
Image 1 is the current BEV map.
Image 2 is the current color BEV map.
Images 3 to 6 are recent first-person views related to the current goal.
The robot has been stagnant near this goal.
The target question is: \{goal\_object\}
If any first-person image contains enough visual evidence to answer the target question, include the concise answer as \texttt{[EQA]answer[/EQA]}.
When visible, the same current goal is marked in the first-person views by a red circular marker labeled 1.
Reply 1 if this goal region should be banned, otherwise reply 0.
The goal is a red point on the BEV and colored BEV map.
Large green circle near the red point indicates that the robot fails to approach the goal point.
If the red-point goal seems irrelevant to the target question or clearly unreachable, ban the goal.
Please follow the format: \\
\texttt{<thought> [reasoning content] </thought>} \\
\texttt{<answer> [1 or 0] </answer>} \\
\texttt{[EQA][only if answerable, concise QA answer][/EQA]}
\end{vlmprompt}

\begin{vlmprompt}[Floor Summary PoI Confirmation]
\textbf{User:} You are helping an indoor robot decide whether it should leave the current floor and search via stairs.
Images 1 to 6 are first-person views selected to summarize covered observations on the current floor.
The target question is: \{goal\_object\}
If any image contains enough visual evidence to answer the target question, include the concise answer as \texttt{[EQA]answer[/EQA]}.
Reply 1 if the robot should go search on another floor now, otherwise reply 0.
Observations leading to dark regions or dark hallways which lacks data don't represent the environment undiscovered.
Even a short flight of stairs can create a barrier between different levels.
Please follow the format: \\
\texttt{<thought> [reasoning content] </thought>} \\
\texttt{<answer> [1 or 0] </answer>} \\
\texttt{[EQA][only if answerable, concise QA answer][/EQA]}
\end{vlmprompt}

\subsection{RLVR Training Hyper-Parameters}

\label{rlvr_training_details}

We use \texttt{algorithm.adv\_estimator=grpo} in the open-source framework \href{https://github.com/hiyouga/EasyR1}{EasyR1} for training. Table~\ref{tab:rlvr_hyperparameters} shows the default hyper-parameters used in the training process.

\begin{table}[htbp]
    \centering
    \caption{Summary of hyper-parameters for trajectory collection and RLVR training.}
    \label{tab:rlvr_hyperparameters}
    \begin{tabular}{@{}llcc@{}}
        \toprule
        \textbf{Category} & \textbf{Parameter Description} & \textbf{Symbol} & \textbf{Value} \\
        \midrule
        \textbf{Sampling} 
        & Dynamic exploration rate & $\epsilon$ & $1 - \min(0.6, 3/d_{goal})$ \\
        & Max greedy probability & $1-\epsilon_{min}$ & 0.6 \\
        & Rollout batch size (prompts per rollout) & $B_{rollout}$ & 256 \\
        & Group size (responses per prompt) & $n$ & 5 \\
        & Sampling temperature & $T$ & 1.0 \\
        & Sampling top-$p$ & $p$ & 1.0 \\
        \midrule
        \textbf{Optimization} 
        & Actor learning rate & $\eta$ & $1.0 \times 10^{-6}$ \\
        & Actor global batch size & $B_{actor}$ & 128 \\
        & Maximum training steps & $T_{max}$ & 50 \\
        & KL penalty type & - & low\_var\_kl \\
        & KL penalty coefficient & $\beta_{\text{KL}}$ & 0.01 \\
        \bottomrule
    \end{tabular}
\end{table}

\subsection{RLVR Training Curves.}
\begin{figure}[t]
\centering
\includegraphics[width=\linewidth]{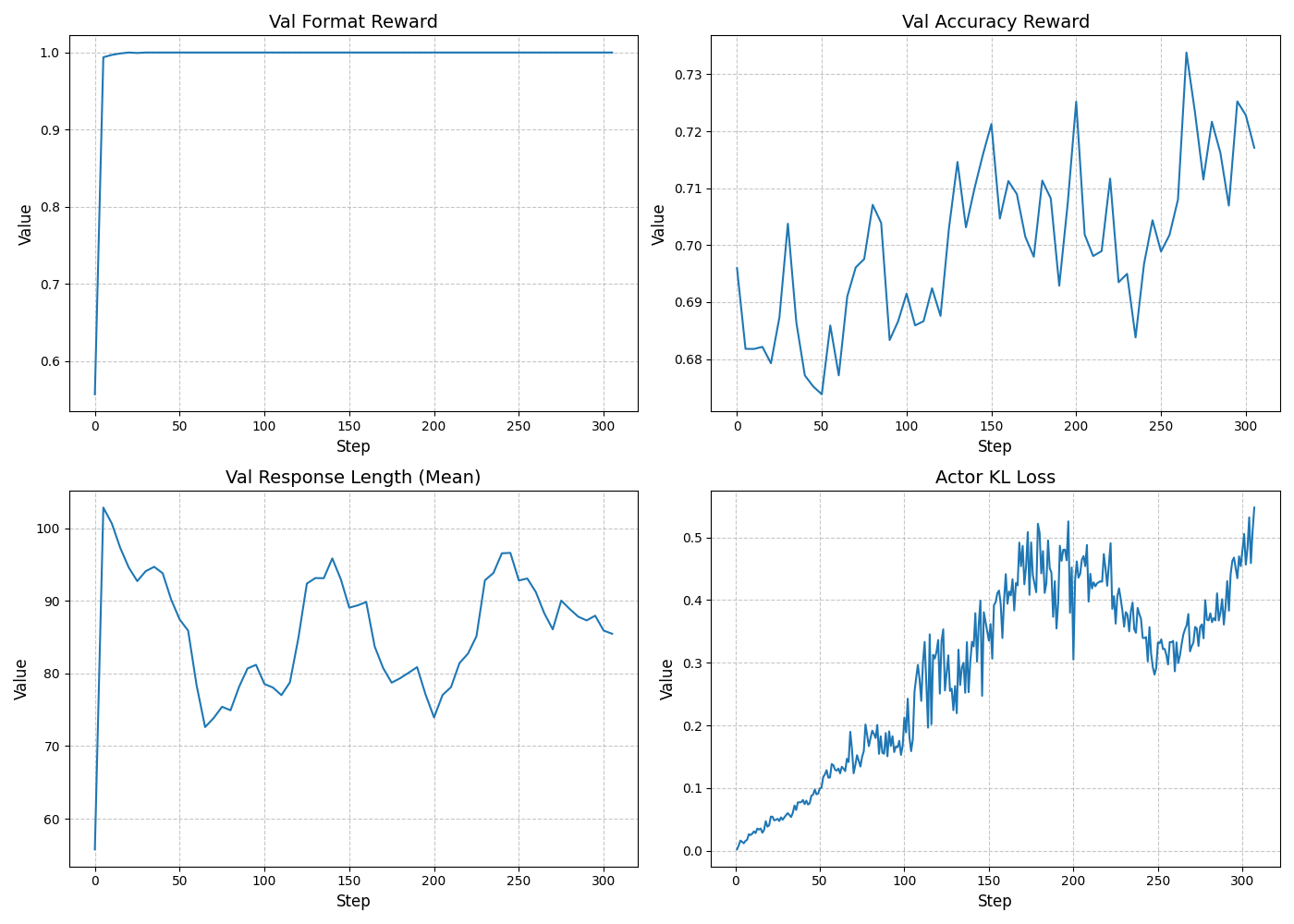} 
\caption{RLVR Training Curves.}
\label{simulation}
\end{figure}

We trained our model using the GRPO algorithm on a hardware setup of 8× H100 GPUs for about 50 hours in our final RLVR vision. The training process utilized all defined rewards and was conducted over 305 steps without a KL penalty. The corpus consisted of 38.7k training questions and 1.62k validation questions. The training dynamics are illustrated in Figure~\ref{simulation}, which tracks the validation rewards, response length, and KL divergence over the course of training.

\textbf{Format and Accuracy Rewards}\quad We observe that the Validation Format Reward converges very rapidly, reaching optimal performance within the first few steps. This indicates that the model quickly adapts to the required structural format. Concurrently, the Validation Accuracy Reward exhibits a steady and continuous upward trend throughout the 305 steps, demonstrating that the model consistently improves its core reasoning and task-solving capabilities.

\textbf{Response Length Dynamics}\quad The rapid convergence of the format reward initially brings a significant boost to the Validation Response Length (Mean). However, as the training progresses, the optimization focuses primarily on the accuracy reward. Consequently, the model's modifications to the response length become less aggressive, transitioning from sharp initial spikes to a more steady and controlled growth pattern.

\textbf{KL Divergence}\quad The Actor KL Loss curve shows a continuous upward trajectory. This illustrates that as the training progresses and the model optimizes for the given rewards, the KL divergence between our trained policy and the original Qwen2.5-VL-7B base model grows increasingly larger.

\subsection{Navigation Hyper-Parameters}
Table~\ref{tab:hyperparameters} shows the hyper-parameters used in the navigation process.
\begin{table}[htbp]
    \centering
    \caption{Summary of hyperparameters used in the system.}
    \label{tab:hyperparameters}
    \begin{tabular}{@{}llcc@{}}
        \toprule
        \textbf{Category} & \textbf{Parameter Description} & \textbf{Symbol} & \textbf{Value} \\
        \midrule
        \textbf{Detection} 
        & Staircase recognition threshold & $\tau_{stair}$ & 0.15 \\
        & Suspect object confidence threshold & $\tau_{sus}$ & 0.25 \\
        & Target object confidence threshold & $\tau_{target}$ & 0.70 \\
        & Max total VLM confirmations & $t_{con}$ & 5 \\
        & Max active VLM confirmations & $t_{acon}$ & 3 \\
        \midrule
        \textbf{Exploration} 
        & Height threshold for sinking regions & $h_{floor}$ & -0.4 m \\
        & Exploration Degree threshold & $\alpha_{exp}$ & 100 \\
        & Stuck Degree threshold & $\alpha_{stuck}$ & 0.01 \\
        & Observations gathered on current floor & $n_{floor}$ & 6 \\
        & Step limit for stuck condition & $t_{stuck}$ & 40 \\
        & Top-ranked stair boundaries & $n_{stair}$ & 4 \\
        & Nearest proximity PoIs collected & $n_{prox}$ & 4 \\
        & Impassable zone radius & $r_{ban}$ & 3.0 m \\
        \bottomrule
    \end{tabular}
\end{table}

\end{document}